%% file: main.tex
\documentclass[10pt,twocolumn,letterpaper]{article}

\usepackage{cvpr}              

\input{preamble}

\definecolor{cvprblue}{rgb}{0.21,0.49,0.74}
\usepackage[pagebackref,breaklinks,colorlinks,allcolors=cvprblue]{hyperref}
\usepackage{caption}
\usepackage{subcaption}

\usepackage{graphicx}
\usepackage{amsmath}
\usepackage{amssymb}
\usepackage{booktabs}
\usepackage{multirow}
\usepackage{pifont} 
\usepackage{adjustbox}
\usepackage{xcolor}
\usepackage{pifont}
\usepackage{xcolor,colortbl}

\usepackage{booktabs}       
\usepackage{amsfonts}       
\usepackage{nicefrac}       
\usepackage{microtype}      
\usepackage{xcolor}         
\usepackage{colortbl}
\usepackage{amsmath}
\usepackage{tikz}
\usepackage{comment}
\usepackage{color}

\usepackage{enumitem}
\usepackage{graphicx}
\usepackage{amsmath}
\usepackage{booktabs}
\usepackage{mathtools}
\usepackage{url}
\usepackage{comment}
\usepackage{color}
\usepackage{graphicx,setspace}
\usepackage{multirow, array, booktabs,makecell,xspace,bm}
\usepackage{pifont}
\usepackage{amsfonts}
\usepackage{arydshln}
\usepackage{dsfont}
\usepackage{stackengine}
\usepackage{scalerel}
\usepackage{esdiff} 
\usepackage{bbm}
\usepackage{algorithm}
\usepackage{algpseudocode}

\def\paperID{38107} 
\def\confName{CVPR}
\def\confYear{2026}

\title{MonoSAOD: Monocular 3D Object Detection with Sparsely Annotated Label}

\author{
Junyoung Jung\thanks{Equal contribution} \qquad
Seokwon Kim\footnotemark[1] \qquad
Jung Uk Kim\thanks{Corresponding author}\\
Kyung Hee University\\
{\tt\small \{jun3700, kimsw0683, ju.kim\}@khu.ac.kr}
}

\begin{document}
\maketitle
\input{sec/main}
{
    \small
    \bibliographystyle{ieeenat_fullname}
    \bibliography{main}
}

\input{sec/X_suppl}

\end{document}

%% file: preamble.tex
\usepackage{graphicx}
\usepackage{multirow}
\usepackage{multicol}
\usepackage{booktabs,multirow}
\usepackage{pifont}
\newcommand{\cmark}{\ding{51}} 
\usepackage{arydshln}

%% file: sec/main.tex

\begin{abstract}
Monocular 3D object detection has achieved impressive performance on densely annotated datasets. 
However, it struggles when only a fraction of objects are labeled due to the high cost of 3D annotation. This sparsely annotated setting is common in real-world scenarios where annotating every object is impractical.
To address this, we propose a novel framework for sparsely annotated monocular 3D object detection with two key modules.
First, we propose Road-Aware Patch Augmentation (RAPA), which leverages sparse annotations by augmenting segmented object patches onto road regions while preserving 3D geometric consistency. 
Second, we propose Prototype-Based Filtering (PBF), which generates high-quality pseudo-labels by filtering predictions through prototype similarity and depth uncertainty. It maintains global 2D RoI feature prototypes and selects pseudo-labels that are both feature-consistent with learned prototypes and have reliable depth estimates.
Our training strategy combines geometry-preserving augmentation with prototype-guided pseudo-labeling to achieve robust detection under sparse supervision.
Extensive experiments demonstrate the effectiveness of the proposed method. The source code is available at \url{https://github.com/VisualAIKHU/MonoSAOD}.
\end{abstract}


\section{Introduction}
\label{sec:intro}

Monocular 3D object detection aims to detect 3D objects using only a single camera, offering cost-effectiveness compared to LiDAR-based methods~\cite{pointcloud1, pointcloud2, pointcloud3} that require expensive sensors. Due to this characteristic, monocular 3D object detection is applied to a wide range of real-world applications such as autonomous driving~\cite{autonomous,autonomous2} and robotics~\cite{robot, robot2}.

\begin{figure}
\centerline{\includegraphics[width=1.0\columnwidth]{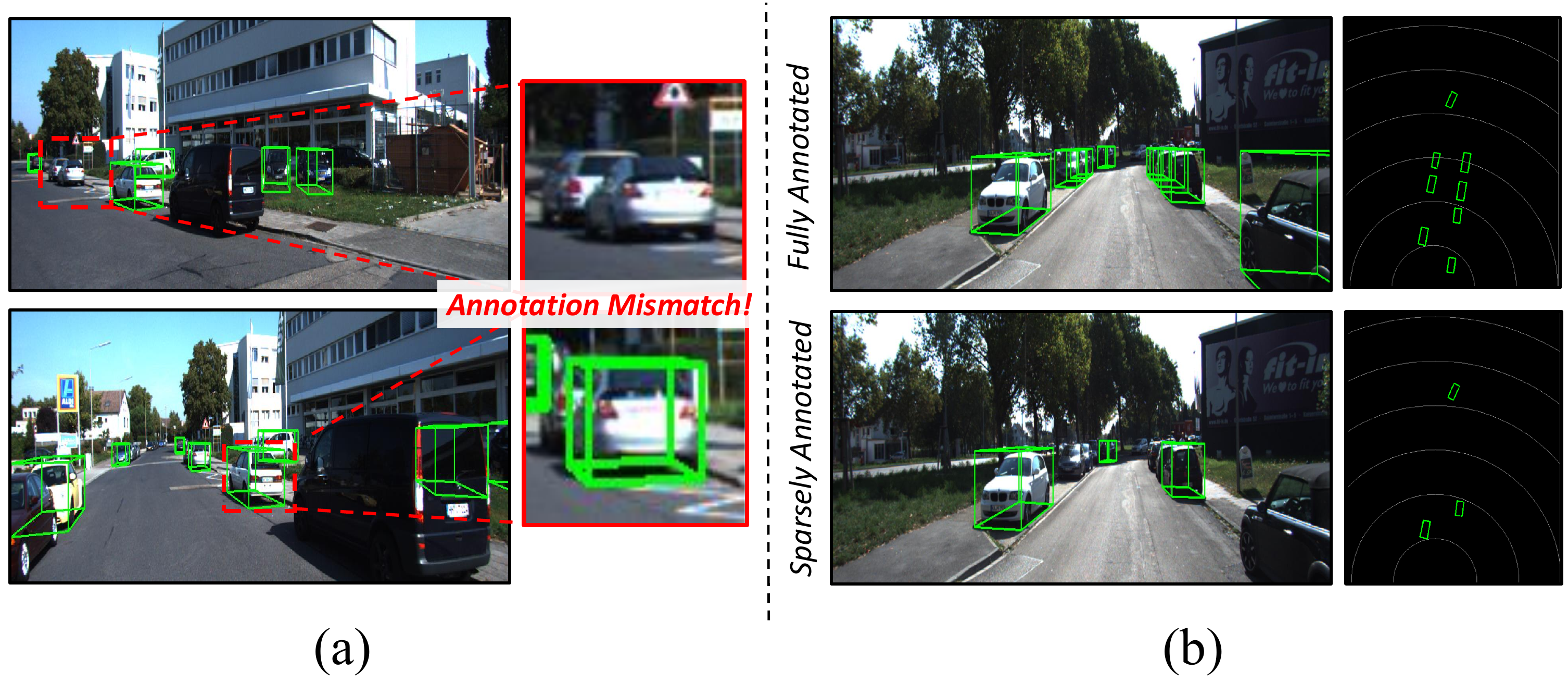}}
\caption{(a) Visible objects are annotated in one scene but missed in another due to the difficulty of 3D bounding box annotation and human error, resulting in inconsistent supervision. (b) Comparison between fully annotated and sparsely annotated labels. In the sparse annotation setting, only a subset of objects is labeled while many valid objects remain unlabeled.}
\label{fig1}
\vspace{-0.2cm}
\end{figure}

While recent approaches~\cite{gupnet, monodtr, monodetr, monocd, monodgp} have made remarkable progress in monocular 3D detection, they typically rely on a highly restrictive assumption that complete 3D annotations are available for all objects. However, obtaining such annotations is highly expensive—it requires precise depth, dimension, and orientation labels—and takes about 3-16 times longer than obtaining 2D annotations~\cite{tang2019transferable}. Consequently, real-world datasets often exhibit \textit{sparse} and \textit{inconsistent} annotations, where some visible objects are missing across images (see Figure~\ref{fig1}(a)). These inconsistencies make it difficult for the model to learn reliable depth and orientation cues, which are crucial for accurate 3D estimation. This results in substantial performance degradation under sparse annotation (see Figure \ref{fig1}(b)) settings.

Recently, to address the above issue, Sparsely Annotated Object Detection (SAOD) task has been actively studied~\cite{comining,calibratedteacher, sparsedet, costudent, ss3d, coin, sp3d, lss3d}. However, the existing SAOD methods are not directly applicable to monocular 3D object detection, which must infer 3D structures solely from 2D images without explicit geometric supervision. Point cloud-based approaches~\cite{ss3d, coin, sp3d, lss3d} exploit LiDAR depth information to generate reliable pseudo-labels, but such geometric cues are absent in monocular settings, making their strategies inapplicable. Likewise, 2D SAOD frameworks~\cite{calibratedteacher, costudent} select pseudo-labels according to confidence scores, which primarily reflect model certainty in 2D localization and classification rather than the accuracy of 3D properties. Consequently, even high-confidence predictions may contain large 3D errors that mislead the model in sparse annotation settings.

In this paper, we propose a novel framework designed to address the challenges of monocular 3D object detection under sparse annotation settings. To this end, we tackle two main challenges: (\textit{i}) enhancing road-object understanding and scene diversity despite the limited availability of ground-truth annotations, and (\textit{ii}) generating reliable pseudo-labels by jointly validating 2D appearance cues and 3D geometric accuracy, enabling effective discrimination between true objects and background predictions.

To address the challenge (\textit{i}), we propose a Road-Aware Patch Augmentation (RAPA) module, which exploits sparse ground truths through realistic data augmentation. The RAPA module strengthens object representation by enabling the model to learn from labeled object instances observed in other scenes. To obtain clean object regions, we use the Segment Anything Model (SAM~\cite{SAM}) to remove surrounding background, after which the isolated objects are reintroduced into the current image at plausible road locations where a vehicle could physically appear. During this process, we account for the 3D geometric changes that naturally occur when an object is repositioned in space, ensuring that the augmented samples remain physically and geometrically consistent. This physically grounded augmentation allows the model to acquire richer and more reliable object representations even under sparse annotations.

For the challenge (\textit{ii}), we introduce a Prototype-Based Filtering (PBF) module, which generates high-quality pseudo-labels by filtering predictions through prototype similarity and depth uncertainty. The PBF module maintains class-specific prototypes that encode canonical appearance patterns learned from reliable detections, enabling it to assess whether new predictions exhibit consistent visual characteristics. By combining feature-level similarity matching with depth uncertainty filtering, the PBF module ensures that selected pseudo-labels have both consistent 2D features and reliable 3D depth estimates. This prototype-guided filtering combined with depth uncertainty effectively rejects predictions with erroneous properties that confidence scores alone fail to detect.

Our main contributions are as follows:
\begin{itemize}
    \item To the best of our knowledge, we are the first to formulate sparsely annotated monocular 3D object detection and reveal the significant performance degradation of existing methods under sparse supervision.
    \item We propose the RAPA module, which exploits sparse annotations by augmenting object patches onto road regions while preserving 3D geometric consistency. 
    \item We introduce the PBF module, which validates both 2D feature consistency and 3D depth reliability to generate high-quality pseudo-labels.
    \item Extensive experiments on KITTI demonstrate substantial improvements under 30\% sparse annotation, outperforming existing approaches.
\end{itemize}


\begin{figure*}[t]
  \centering
  \includegraphics[width=0.99\textwidth]{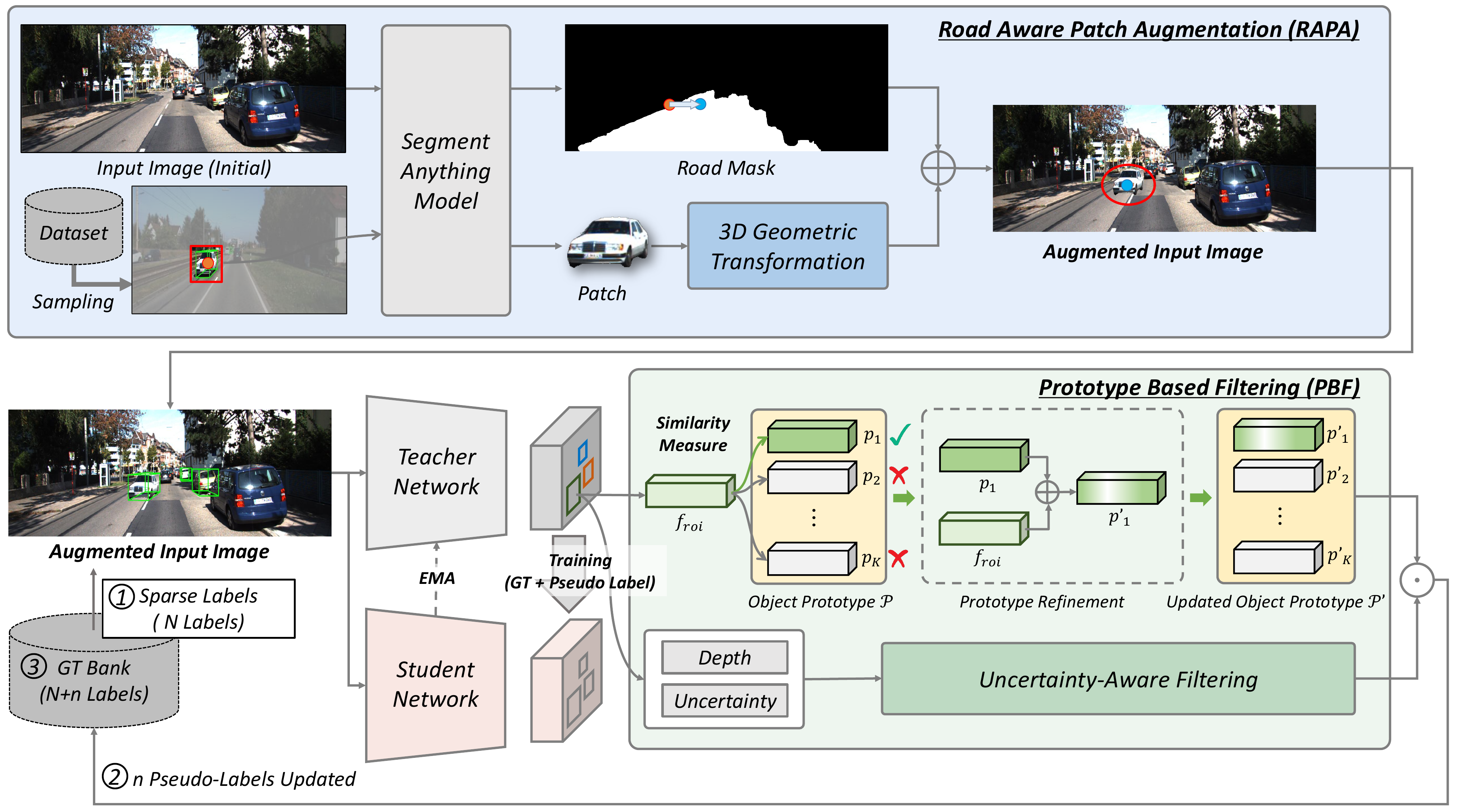}
  \caption{Overall architecture of the proposed framework.
  A teacher–student structure integrates Road-Aware Patch Augmentation (RAPA) and Prototype-Based Filtering (PBF) for robust training under sparse annotations.}
  \label{fig:overview}
\vspace{-0.3cm}
\end{figure*}

\section{Related Work}
\label{sec:related}

\subsection{Monocular 3D Object Detection}

Monocular 3D object detection (M3OD) aims to detect objects in three-dimensional space using only a single camera. Existing methods address the core challenge of reconstructing spatial information from 2D images through various approaches. Geometry-based methods~\cite{deep3dbox, m3d-rpn, monoflex, monocd} utilize geometric constraints between 2D and 3D spaces to infer depth information. Pseudo-LiDAR approaches~\cite{pseudo-lidar1, pseduo-lidar2, pseudo-lidar3, pseudo-lidar4} convert images into pseudo point clouds using depth estimation, enabling the use of LiDAR-based detectors. Other depth-assisted methods~\cite{caddn, kim2023stereoscopic} explicitly incorporate depth cues into the detection pipeline. Recently, transformer-based methods~\cite{monodtr, monodetr, monodgp} leverage attention mechanisms for effective feature aggregation. Separately, Omni3D~\cite{brazil2023omni3d} introduces a large-scale multi-dataset benchmark and trains a unified detector across diverse domains, while DetAny3D~\cite{zhang2025detect} enables promptable zero-shot 3D detection by leveraging priors from 2D foundation models.

While these methods have shown remarkable performance, they rely on dense and accurate 3D annotations, leaving the sparse annotation regime unexplored. Obtaining such precise annotations is labor-intensive and costly, and real-world datasets often face annotation limitations due to occlusions and labeling difficulties. This motivates our work to explore M3OD in sparsely annotated settings, where only a fraction of objects in training images are labeled.

\subsection{Sparsely Annotated Object Detection}
Sparsely annotated object detection (SAOD) addresses the challenge of training detectors when only a subset of instances are labeled in each image. This setting differs fundamentally from semi-supervised object detection (SSOD), where some images are fully labeled and others are entirely unlabeled. Early SAOD methods mainly focused on reducing the negative effects of unlabeled instances by reweighting their gradients or recalibrating the background losses. While these strategies mitigate the issue of missing annotations, they do not actively exploit unlabeled regions for learning.

Recent works have shifted toward mining pseudo-labels from unlabeled instances to provide additional supervision. Co-mining~\cite{comining} adopts a Siamese architecture with weak and strong augmentations to generate complementary pseudo-labels between two branches. SparseDet~\cite{sparsedet} separates proposals into labeled, unlabeled, and background regions and applies self-supervised consistency loss to unlabeled regions. Calibrated Teacher~\cite{calibratedteacher} addresses the instability of pseudo-label thresholds by calibrating confidence scores of the teacher to reflect its true precision, enabling consistent pseudo-label selection across different training stages.

However, these SAOD methods rely primarily on classification confidence scores to select pseudo-labels, which can be unreliable especially in monocular 3D object detection, since each prediction includes complex geometric attributes beyond class labels. In contrast, we introduce Prototype-Based Filtering (PBF), which evaluates pseudo-labels based on their feature similarity to learned prototypes and their estimated depth uncertainty, providing a more reliable 
measure of pseudo-label quality.

\subsection{Data Augmentation for Object Detection}

Data augmentation has proven effective for improving object detection performance by increasing training data diversity. Traditional augmentation strategies include geometric transformations such as random flipping, scaling, and rotation, as well as photometric variations. For 3D object detection, GT-sampling~\cite{ss3d} has become a standard technique, while copy-paste augmentation~\cite{sampd} has shown promising results in image-based 2D object detection. Recent semi-supervised monocular 3D object detection methods~\cite{mix-teaching, alleviatingWACV} adopt patch-based augmentation by copying and pasting image patches from labeled data.
However, these methods typically rely on 2D bounding box coordinates, inevitably including background regions that lead to visually unrealistic composites and may introduce noise during training. 
Moreover, they lack spatial constraints such as road boundaries, often resulting in implausible object placements.

To overcome these limitations, we propose the Road-Aware Patch Augmentation (RAPA) module that exploits SAM~\cite{SAM} for precise instance segmentation, extracting only the object region, and employs road segmentation masks to guide patch placement. Additionally, we enable objects to be spatially repositioned while preserving their 3D coordinates through geometric transformations that maintain depth and orientation consistency. It ensures that augmented objects remain geometrically plausible in 3D space while being placed at realistic locations within the scene.



\section{Proposed Methods}

Figure~\ref{fig:overview} shows the overall framework of our method. The input image and sparse annotations are processed through a teacher-student architecture with two key components: Road-Aware Patch Augmentation (RAPA) and Prototype-Based Filtering (PBF). Using SAM~\cite{SAM}, the RAPA module segments object patches from other scenes and extracts road masks of the target image. These patches are geometrically transformed and pasted onto valid road regions while maintaining realistic scale, depth, and orientation to generate augmented images. The teacher network processes these augmented images to produce 3D predictions. Next, given the teacher predictions, the PBF module evaluates their reliability by jointly considering 2D semantic attributes (feature-level consistency) through prototype similarity and 3D geometric attributes (depth reliability) through uncertainty estimation. Selected pseudo-labels are then utilized in two ways: they update the prototypes through cumulative aggregation to enrich object representations, and reliable pseudo-labels are stored in a GT Bank to serve as ground-truth annotations in subsequent epochs. Finally, the student network is trained on both sparse ground-truth labels and the PBF module filtered pseudo-labels. More details are provided in the following subsections.

\subsection{Road-Aware Patch Augmentation (RAPA)}
Under sparse annotation settings, each labeled object must be leveraged fully to compensate for the scarcity of training data. However, existing copy-paste augmentation strategies~\cite{mix-teaching, sampd, alleviatingWACV} for object detection produce unrealistic training samples. These methods typically paste entire rectangular regions including background context, leading to unnatural artifacts. Moreover, they place augmented objects at their original scene locations without translation, resulting in objects floating above the ground or appearing in invalid locations such as sidewalks or buildings. To address these limitations, we introduce the Road-Aware Patch Augmentation (RAPA) module, a data augmentation strategy that synthesizes realistic training samples by placing cleanly segmented car instances from other scenes into geometrically valid locations while maintaining 3D consistency. \\

\noindent\textbf{Patch Extraction.} Given the training images, we first extract high-quality object patches (\textit{i.e.,} objects without truncation and occlusion) from the sparse ground-truth annotations. For each extracted object patch, we adopt SAM~\cite{SAM} using its 2D bounding box as a prompt to segment the object from the background. We also generate road segmentation masks $M_{road}$ for each scene using SAM with manually provided point prompts on the road regions. \\

\noindent\textbf{3D-Aware Transformation.} After extracting the object patches, the key challenge is preserving the visual appearance of the object while correctly transforming its 3D pose. Given a patch $P_s$ from the source image $I_s$ to be augmented onto the target image $I_t$, we first transform the 3D center of $P_s$ from source to target camera coordinates using extrinsic matrix:
\begin{align}
\begin{bmatrix}
x_t \\ y_t \\ z_t
\end{bmatrix}
=
[R_t \mid T_t]\,[R_s \mid T_s]^{-1} \,
\begin{bmatrix}
x_s \\ y_s \\ z_s
\end{bmatrix},
\end{align}
where $[R_s \mid T_s]^{-1}$ denotes the inverse extrinsic matrix.\\

\noindent\textbf{Horizontal Translation.} We explore candidate positions by applying horizontal offsets to the transformed 3D center:
\begin{align}
\begin{bmatrix}
x_t' \\ y_t' \\ z_t'
\end{bmatrix}
=
\begin{bmatrix}
x_t \\ y_t \\ z_t
\end{bmatrix}
+
\begin{bmatrix}
x_{\text{offset}} \\ 0 \\ 0
\end{bmatrix},
\end{align}
where $x_{\text{offset}} \in [-\delta, \delta]$ with $\delta = s$ meters, sampled at $m$ uniform intervals. \\

\noindent\textbf{Rotation Update.} For each candidate position, we update the orientation of the object to maintain its visual appearance at the new viewpoint. We preserve the observation angle $\alpha$ (which describes how the object appears relative to the camera) while computing the new rotation $r'_y$:
\begin{align}
\theta_{\text{new}} &= \arctan2(x_t', z_t'), \\[6pt]
r_y' &= \alpha + \theta_{\text{new}},
\end{align}
where $\theta_{\text{new}}$ is the viewing angle at the new position. This ensures that an object that appears front-facing in the source image will still appear front-facing after augmentation, regardless of its new position. Importantly, the physical dimensions of the object $(h, w, l)$ remain unchanged during transformation. \\

\noindent\textbf{Validity Check.} We then project the full 3D box $(x'_t,y'_t,z'_t,h,w,l,r'_y)$ to the 2D image plane and enforce two constraints. First, we ensure the augmented object appears on drivable surfaces:
\begin{equation}
\frac{\sum_{(i,j) \in \mathbf{b}} \mathbf{M}_{\text{road}}(i,j)}{|\mathbf{b}|} \geq \tau_{\text{road}},
\end{equation}
where \textbf{b} is the projected 2D bounding box and $\tau_{\text{road}}$ is a threshold. Second, we prevent unrealistic overlap with existing labeled objects:
\begin{equation}
\mathrm{IoU}(\mathbf{b}_{\text{new}}, \mathbf{b}_e) < \tau_{\text{overlap}}, \quad \forall \mathbf{b}_e \in \mathcal{B}_{\text{existing}},
\end{equation}
where $\mathcal{B}_{existing}$ contains all existing bounding boxes in that scene. Once a valid position is found, we resize the patch to match the projected 2D bounding box size and place it onto the target image using the segmentation mask.

By explicitly handling 3D geometry through camera calibration and preserving visual consistency through observation angle, the RAPA module effectively leverages sparse annotations while ensuring that augmented samples remain visually and geometrically valid for monocular 3D object detection training. \\

\subsection{Prototype-Based Filtering (PBF)}

While existing 2D SAOD methods select pseudo-labels based on classification confidence scores, this is insufficient in monocular 3D object detection: confidence scores reflect 2D localization certainty but not the accuracy of 3D properties such as depth and orientation. As a result, a prediction may exhibit high confidence while containing large 3D errors that severely mislead the student network, particularly at higher sparsity levels where the teacher has limited supervision to learn accurate 3D estimation. To address this, we introduce the Prototype-Based Filtering (PBF) module, which evaluates pseudo-label reliability by jointly considering 2D semantic attributes (feature-level consistency) and 3D geometric attributes (depth reliability), ensuring that selected pseudo-labels are both semantically and geometrically trustworthy, which is crucial for learning accurate 3D object representations under sparse supervision. \\

\noindent\textbf{Prototype Initialization.}
Before training begins, we initialize the prototypes $\mathcal{P} = \{\, p_k \,\}_{k=1}^{K}$
using sparse ground-truth labels ($K$ is the number of the capacity). This pre-computation establishes a reliable reference distribution of object Region-of-Interest (RoI) features that guides subsequent pseudo-label selection. For each ground-truth object in the training set, we extract RoI features $f_{roi}$ from the teacher network.

During initialization, features with high cosine similarity (above threshold $\tau_{\text{new}} = 0.8$) to existing prototypes are merged through a weighted cumulative update:
\begin{equation}
p_k' = (1 - \beta)\, p_k + \beta\, f_{\text{roi}},
\label{cumulative_sum}
\end{equation}
where $\beta$ is the update weight. Distinct features form new prototypes until the bank reaches its capacity $K$. This initialization process ensures that the prototypes capture diverse and reliable object representations from the limited ground-truth annotations, providing a robust foundation for filtering pseudo-labels during training. Through this process, all updated prototypes form the refined prototype set $\mathcal{P}'$, which is later used for pseudo-label filtering.\\

\noindent\textbf{Geometric Reliability Filtering.}
To assess the reliability of the predicted depth ${d}_{pred}$, we utilize the predicted uncertainty $\sigma$ derived from the Laplacian aleatoric uncertainty loss~\cite{kendall2017uncertainties, chen2020monopair}:
\begin{equation}
\mathcal{L}_{\text{depth}} =
\frac{\sqrt{2}}{\sigma} \left\lVert d_{\text{gt}} - d_{\text{pred}} \right\rVert_1
+ \log(\sigma),
\label{eq:depth_loss}
\end{equation}
where $\sigma$ denotes the predicted uncertainty. This formulation encourages the network to assign small $\sigma$ values for confident predictions and larger $\sigma$ values for ambiguous cases.

During pseudo-label generation, we define a geometric reliability score as:
\begin{equation}
S_{\text{depth}} = \exp(-\sigma).
\end{equation}
A higher $S_{\text{depth}}$ indicates more reliable geometric predictions. Only predictions with $S_{\text{depth}} > \tau_{\text{depth}}$ proceed to the next filtering stage. \\

\noindent\textbf{Semantic Consistency Filtering.} Let $\{\, f_{\text{roi}}^{(i)} \,\}_{i=1}^{N}$ denote the set of RoI features extracted from $N$ candidate predictions that pass geometric filtering. For each $f_{\text{roi}}^{(i)}$, semantic consistency is evaluated by measuring its maximum cosine similarity to all prototypes in $\mathcal{P}$:
\begin{equation}
S_{\text{proto}}^{(i)} =
\max_{p_k \in \mathcal{P}}
\left(
\frac{
f_{\text{roi}}^{(i)} \cdot p_k
}{
\lVert f_{\text{roi}}^{(i)} \rVert \,
\lVert p_k \rVert
}
\right).
\end{equation}
This score quantifies how well candidate $i$ aligns with the matched prototype feature.\\

\noindent\textbf{Final Pseudo-Label Selection.}
A prediction is selected as a high-quality pseudo-label only when it satisfies both geometric and semantic criteria:
\begin{equation}
S_{\text{depth}} > \tau_{\text{depth}}
\quad \text{and} \quad
S_{\text{proto}} > \tau_{\text{proto}}.
\end{equation}
Only predictions that pass both thresholds are preserved as pseudo-labels for training the student model. This dual-criteria selection ensures that the pseudo-labels maintain reliable depth estimates and semantically consistent features, effectively preventing propagation of erroneous supervision. \\

\begin{table*}[t]
  \caption{Detection results of car category on the KITTI validation set under sparse annotation ratios of 30\%, 50\%, and 70\%. We compare our method with existing Sparsely Annotated Object Detection (SAOD) approaches reproduced using their official implementations for fair comparison. \textbf{Bold}/\underline{underline} fonts indicate the best/second-best results.}
  \renewcommand{\tabcolsep}{3mm}
  \label{tab:sota_comparison}
  \centering
  \resizebox{0.94\linewidth}{!}{
  \begin{tabular}{@{}ccccccccccc@{}}
    \Xhline{3\arrayrulewidth}
    \rule{0pt}{10pt}\multirow{2}{*}{\bf Method} & \multicolumn{3}{c}{\bf 30\%} & \multicolumn{3}{c}{\bf 50\%} & \multicolumn{3}{c}{\bf 70\%}\\ \cmidrule(lr){2-4} \cmidrule(lr){5-7} \cmidrule(lr){8-10}
    & \textbf{Easy} & \textbf{Mod.} & \textbf{Hard}
    & \textbf{Easy} & \textbf{Mod.} & \textbf{Hard}
    & \textbf{Easy} & \textbf{Mod.} & \textbf{Hard} \\\hline
        \rule{0pt}{9.0pt}Baseline~\cite{monodetr}
      & 11.17 & 8.73 & 7.56
      & 20.36 & 15.25 & 12.76
      & 22.78 & 17.83 & 14.97 \\\cdashline{1-10}
        \rule{0pt}{9.0pt}Co-mining~\cite{comining}
      & 16.01 & 12.62 & 10.38
      & 21.47 & 16.22 & \underline{13.61}
      & 24.10 & 18.21 & 15.16 \\

    SparseDet~\cite{sparsedet}
      & 16.95 & \underline{13.30} & \underline{10.97}
      & 21.99 & 16.13 & 13.43
      & 23.58 & 18.39 & 15.54 \\

    Calibrated Teacher~\cite{calibratedteacher}
      & \underline{17.14} & 12.96 & 10.58
      & 22.04 & 16.03 & 13.36
      & \underline{25.39} & \underline{18.94} & \underline{15.86} \\

        Co-student~\cite{costudent}
      & 15.99 & 12.67 & 10.38
      & \underline{23.06} & \underline{16.39} & 13.45
      & 23.71 & 18.11 & 15.12
     \\\cdashline{1-10}
    \rule{0pt}{9.0pt}\textbf{Proposed Method}
      & \textbf{21.28} & \textbf{15.60} & \textbf{12.79}
      & \textbf{26.31} & \textbf{18.84} & \textbf{15.71}
      & \textbf{26.67} & \textbf{19.37} & \textbf{16.25} \\
    \Xhline{3\arrayrulewidth}
  \end{tabular}
  }
  \vspace{-0.1cm}
\end{table*}

\begin{table}[t]
\caption{Detection results of car category on the KITTI test set under 30\% annotation setting. Our method achieves large improvements over existing SAOD methods. \textbf{Bold}/\underline{underline} fonts indicate the best/second-best results.}
\renewcommand{\tabcolsep}{3.2mm}
\label{tab:test_comparison}
\centering
\resizebox{\linewidth}{!}{
\begin{tabular}{c ccc}
\Xhline{3\arrayrulewidth}
\rule{0pt}{9.0pt}\textbf{Method} & \textbf{Easy} & \textbf{Mod.} & \textbf{Hard} \\\hline
\rule{0pt}{9.0pt}Baseline~\cite{monodetr}         & 7.12 & 5.12 & 3.90 \\\cdashline{1-4}
Co-mining~\cite{comining}        & 9.93 & 6.41 & 5.25 \\
SparseDet~\cite{sparsedet}        & \underline{10.76} & \underline{7.42} & \underline{6.16} \\
Calibrated Teacher~\cite{calibratedteacher} & 9.26 & 6.09 & 5.02 \\
Co-student~\cite{costudent}       & 9.96 & 6.97 & 5.33 \\\cdashline{1-4}
\rule{0pt}{9.0pt}\textbf{Proposed Method}       & \bf 17.47 & \bf 11.36 & \bf 8.83 \\
\Xhline{3\arrayrulewidth}
\end{tabular}
}
\end{table}

\noindent\textbf{Prototype Refinement.}
As training progresses, the PBF module continuously refines the prototype bank using the pseudo-labels that have passed both geometric and semantic filtering. After each batch, the RoI features associated with these validated pseudo-labels are merged into their corresponding prototypes via the same weighted cumulative update defined in Eq. \eqref{cumulative_sum}. This ongoing refinement allows the prototype bank to adapt to the evolving feature distribution while maintaining robustness to noisy predictions. \\

\noindent\textbf{GT Bank Update.}
Validated pseudo-labels are also stored in a GT Bank, where they are treated as additional ground-truth annotations in subsequent epochs. This gradually increases the amount of effective training data available to the student model, improving supervision coverage under sparse labels and stabilizing the overall training process.


\begin{table}[t!]
  \caption{Ablation study of the proposed modules on the KITTI validation set for car category under 30\% annotation ratio. `Conf.': confidence score-based pseudo-labeling. `RAPA': Road-Aware Patch Augmentation, `PBF': Prototype-Based Filtering.}
  \renewcommand{\tabcolsep}{3mm}
  \label{tab:ablation1}
  \centering
  \resizebox{\linewidth}{!}{
  \begin{tabular}{cccccc}
    \Xhline{3\arrayrulewidth}
    \textbf{Conf.} &
    \textbf{RAPA}&
    \textbf{PBF} & \textbf{Easy} & \textbf{Mod.} & \textbf{Hard} \\\hline
    - & - &-& 11.17 & 8.73 & 7.56 \\
    \cdashline{1-6}
    \rule{0pt}{9.0pt}\cmark & - &-& 12.39 & 9.68 & 8.18 \\
    \cmark  &- & \cmark  & 16.49 & 12.65 & 10.32 \\  
    \cmark &\cmark & - & 20.31 & 14.51 & 11.72 \\
    \cmark &\cmark &  \cmark & \textbf{21.28} & \textbf{15.60} & \textbf{12.79} \\
    \Xhline{3\arrayrulewidth}
  \end{tabular}
  }
  \vspace{-0.1cm}
\end{table}

\section{Experiments}

\subsection{Experimental Setup}
\noindent\textbf{Dataset and Metrics.} We adopt \textbf{KITTI 3D object detection dataset}~\cite{kitti}. It contains 7,481 training images and 7,518 testing images. Following the common practice~\cite{autonomous2}, the training set is split into 3,712 training samples and 3,769 validation samples. To evaluate performance under more challenging real-world conditions, we additionally adopt \textbf{foggy KITTI dataset}~\cite{monowad}. As in prior 2D SAOD methods~\cite{sparsesampling, sparsedet, comining, sampd}, we randomly select 30\%, 50\%, 70\% of the total annotations. The annotation subsets used in our experiments will be publicly released. We measure performance using average precision for both 3D detection (AP$_{3D}$) and bird’s-eye-view detection (AP$_{BEV}$) under the three standard difficulty levels (`Easy', `Moderate', `Hard'), defined by object size, occlusion, and truncation, and report all results using $AP_{R_{40}}$ metric with IoU threshold 0.7. \\

\noindent\textbf{Implementation Details.} We adopt MonoDETR~\cite{monodetr} and MonoDGP~\cite{monodgp} as our baseline monocular 3D detectors. Among them, we primarily use MonoDETR as the representative model for evaluation, while MonoDGP is additionally employed to assess the generalization ability of our approach. All experiments are conducted on a single RTX 3090 GPU with a batch size of 16. For the backbone, we use a ResNet-50~\cite{he2016deep} for MonoDETR and MonoDGP. We first pre-train the model using sparsely annotated data with the RAPA module to initialize both the teacher and student networks. 
We use AdamW as the optimizer without warm-up and train the model for 100 epochs. For the PBF module, we set the prototype capacity to $K{=}256$, the depth reliability threshold to $\tau_{\text{depth}}{=}1.0$, prototype similarity threshold to $\tau_{\text{proto}}{=}0.85$, and the prototype update weight to $\beta{=}0.005$, while using $\beta{=}0.01$ during prototype initialization. More details are in the supplementary document.


\begin{figure*}[t]
\centerline{\includegraphics[width=0.96\textwidth]{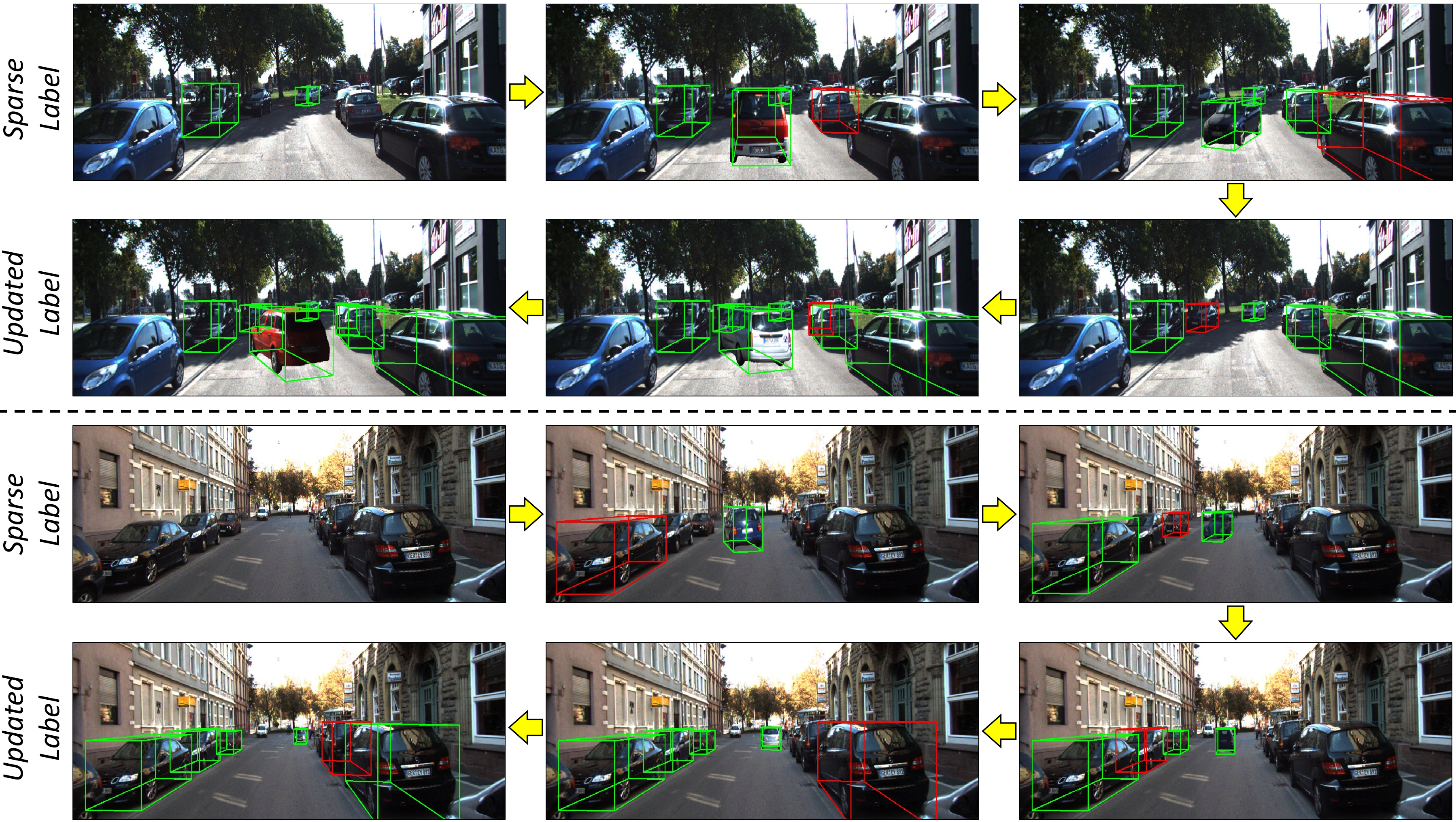}}
\caption{Progression of pseudo-labels selected by the proposed PBF module for GT Bank enrichment. Green boxes denote sparse ground truths and previously accumulated pseudo-labels, while red boxes indicate high-quality pseudo-labels newly selected at the current step. The consistent selection of geometrically and semantically reliable pseudo-labels highlights the effectiveness of the PBF module.}
\label{fig:PBF}
\end{figure*}

\subsection{Comparison with Other Methods}

We compare our method with existing SAOD methods on the KITTI validation set under various annotation percentages (30\%, 50\%, 70\%). Table~\ref{tab:sota_comparison} reports the AP$_{3D}$ results against the baseline, trained solely on sparse annotations, as well as against Co-mining~\cite{comining}, SparseDet~\cite{sparsedet}, Calibrated Teacher~\cite{calibratedteacher}, and Co-student~\cite{costudent}. (More results regarding AP$_{BEV}$ are in the supplementary document.) While all existing methods show improvements over the baseline, our method delivers the highest performance.

We further evaluate our method on the KITTI test set under the 30\% annotation setting. As shown in Table~\ref{tab:test_comparison}, our method achieves superior performance, confirming its effectiveness. We find that existing SAOD methods struggle to preserve reliable 3D information, often producing inaccurate pseudo-labels under sparse annotations. The core challenge lies in assessing the geometric reliability of predicted depth, which conventional confidence measures cannot capture. 
In contrast, we overcome this limitation through the PBF module, which evaluates feature consistency using learned prototypes and verifies depth reliability via uncertainty estimation, enabling more accurate pseudo-label selection for sparsely annotated monocular 3D detection.

\subsection{Ablation Study}

We conduct ablation studies to investigate the effect of our proposed modules. All experiments are conducted on KITTI validation set with 30\% annotation. As shown in Table~\ref{tab:ablation1}, compared to the baseline trained with only sparse ground-truth labels, when confidence-based pseudo-labeling is applied, the improvement is minimal, indicating that relying only on confidence scores derived from 2D information is insufficient for selecting reliable 3D pseudo-labels. Adding either the RAPA module or the PBF module leads to clear performance gains by leveraging 3D-aware representations, surpassing approaches that depend only on 2D confidence. When both modules are integrated, our full method delivers the highest performance, verifying that the RAPA module and the PBF module provide complementary benefits for sparsely annotated monocular 3D object detection.

\begin{figure}
\centerline{\includegraphics[width=\columnwidth]{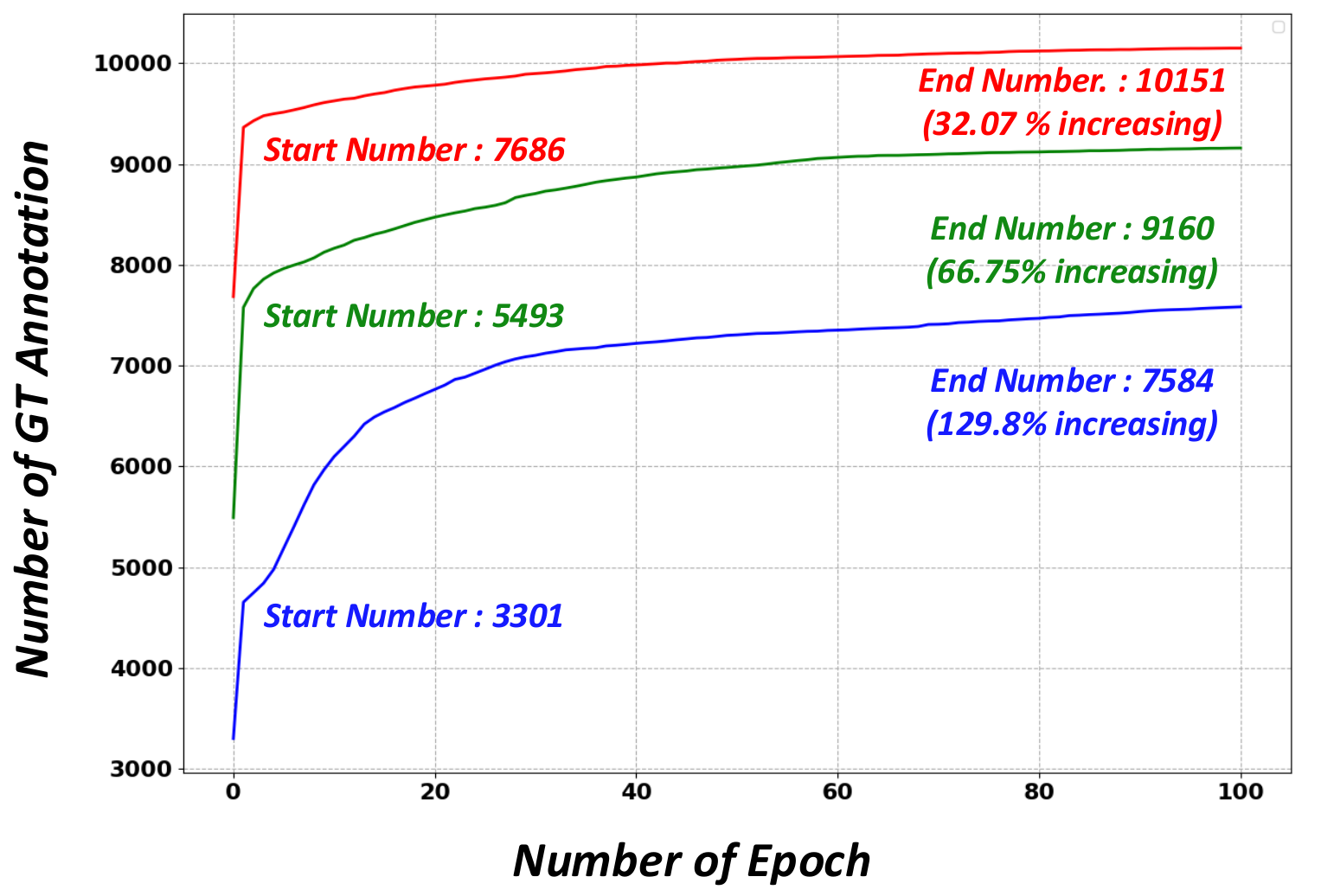}}
\caption{Line plot of GT annotation growth, where 30\% (blue), 50\% (green), and 70\% (red) represent different sparsity levels.}
\label{fig:line_graph}
\end{figure}

\section{Discussion}
\noindent\textbf{Examples of the Progression of Our Method.} Figure~\ref{fig:PBF} and Figure~\ref{fig:line_graph} illustrate how our framework evolves throughout the training process. As shown in Figure~\ref{fig:PBF}, the RAPA module continuously adds clean object patches onto valid road regions while respecting their underlying 3D geometric properties, allowing the detector to observe diverse and geometrically plausible object placements over time. During training, low-quality predictions remain filtered out, while high-quality pseudo-labels—those satisfying both depth and prototype criteria—are gradually promoted from red (candidate) to green (accepted) and stored as ground-truth annotations. Figure~\ref{fig:line_graph} quantitatively shows this progression, highlighting the steady growth of accumulated pseudo-labels in the GT Bank. Overall, the visualizations demonstrate the synergy between the RAPA and PBF modules: the RAPA module enriches the input distribution with realistic placements, and the PBF module progressively expands reliable supervision by selecting trustworthy pseudo-labels. \\

\begin{table}[t]
  \caption{Generalization results of our framework applied as a plug-in to different monocular 3D detectors on the KITTI validation set(30\% annotation). Incorporating our method yields consistent improvements across architectures.}
  \renewcommand{\tabcolsep}{2.4mm}
  \label{tab:plugin_results}
  \centering
  \resizebox{\linewidth}{!}{
  \begin{tabular}{c ccc}
    \Xhline{3\arrayrulewidth}
    \rule{0pt}{9.5pt}\textbf{Method} & \textbf{Easy} & \textbf{Mod.} & \textbf{Hard}\\\hline
    \rule{0pt}{9.0pt}MonoDETR (ICCV'23)~\cite{monodetr} & 11.17 & 8.73 & 7.56  \\
    \textbf{MonoDETR + Ours}& \textbf{21.28} & \textbf{15.60} & \textbf{12.79} \\\cdashline{1-4}
    \rule{0pt}{9.0pt}MonoDGP (CVPR'25)~\cite{monodgp} & 15.56 & 11.70 & 9.53  \\
    \textbf{MonoDGP + Ours} &\textbf{22.62} & \textbf{16.79} & \textbf{14.07} \\ 
    \Xhline{3\arrayrulewidth}
  \end{tabular}
  }
\end{table}

\begin{table}[t!]
\centering
\caption{Performance under foggy KITTI (fog density 0.1) validation set, evaluated under the 30\% annotation setting. Our method demonstrates improved robustness in adverse weather conditions compared to existing SAOD methods.}
\renewcommand{\tabcolsep}{3.2mm}
\label{tab:fog}
\resizebox{\linewidth}{!}{
    \begin{tabular}{cccc}
    \Xhline{3\arrayrulewidth}
    \textbf{Method} & \textbf{Easy} & \textbf{Mod.} & \textbf{Hard} \\
    \hline
    \rule{0pt}{9.5pt}Baseline   & 11.10 & 7.43 & 5.81 \\\cdashline{1-4}
    Co-mining~\cite{comining}  & 11.22 & 7.91 & 6.20 \\
    Sparsedet~\cite{sparsedet}  & 11.40 & 7.93 & 6.52 \\
    Calibrated Teacher~\cite{calibratedteacher} & \underline{11.97} & \underline{8.65} & \underline{7.26} \\
    Co-student~\cite{costudent} & 11.57 & 7.97 & 6.20
    \\\cdashline{1-4}
    \rule{0pt}{9.0pt}\textbf{Proposed Method }       & \textbf{19.11} & \textbf{13.72} & \textbf{10.35} \\
    \Xhline{3\arrayrulewidth}
    \end{tabular}
}
\vspace{-0.2cm}
\end{table}

\noindent\textbf{Generalization Ability.} To evaluate the generalization ability of our method, we apply our framework (\textit{i.e.,} RAPA and PBF modules) in a plug-in manner to the recent monocular 3D detector, MonoDGP~\cite{monodgp}. As shown in Table~\ref{tab:plugin_results}, our approach consistently shows substantial performance improvements on the KITTI validation set under 30\% annotation setting. These results demonstrate that both RAPA and PBF modules can be effectively integrated into various monocular 3D detection architectures, providing reliable gains regardless of the underlying model design. \\

\noindent\textbf{Results on Foggy Scenarios.} To investigate the robustness under challenging conditions, we conduct additional experiments on the foggy KITTI dataset~\cite{monowad} (fog density$=$0.1). From a labeling perspective, fog makes accurate annotation difficult because objects become partially invisible and depth cues are degraded, leading to more missing or unreliable labels. This amplifies the challenges of sparse annotation, making foggy KITTI a meaningful benchmark for evaluating the robustness.

As shown in Table~\ref{tab:fog}, our method demonstrates noticeably stronger robustness compared to the baselines. The RAPA module improves feature learning by introducing diverse object appearances, while the PBF module further enhances reliability by filtering pseudo-labels through feature consistency and depth uncertainty. Overall, our approach effectively addresses both sparse annotation challenges and adverse weather conditions. \\

\begin{figure}
\centerline{\includegraphics[width=1.02\columnwidth]{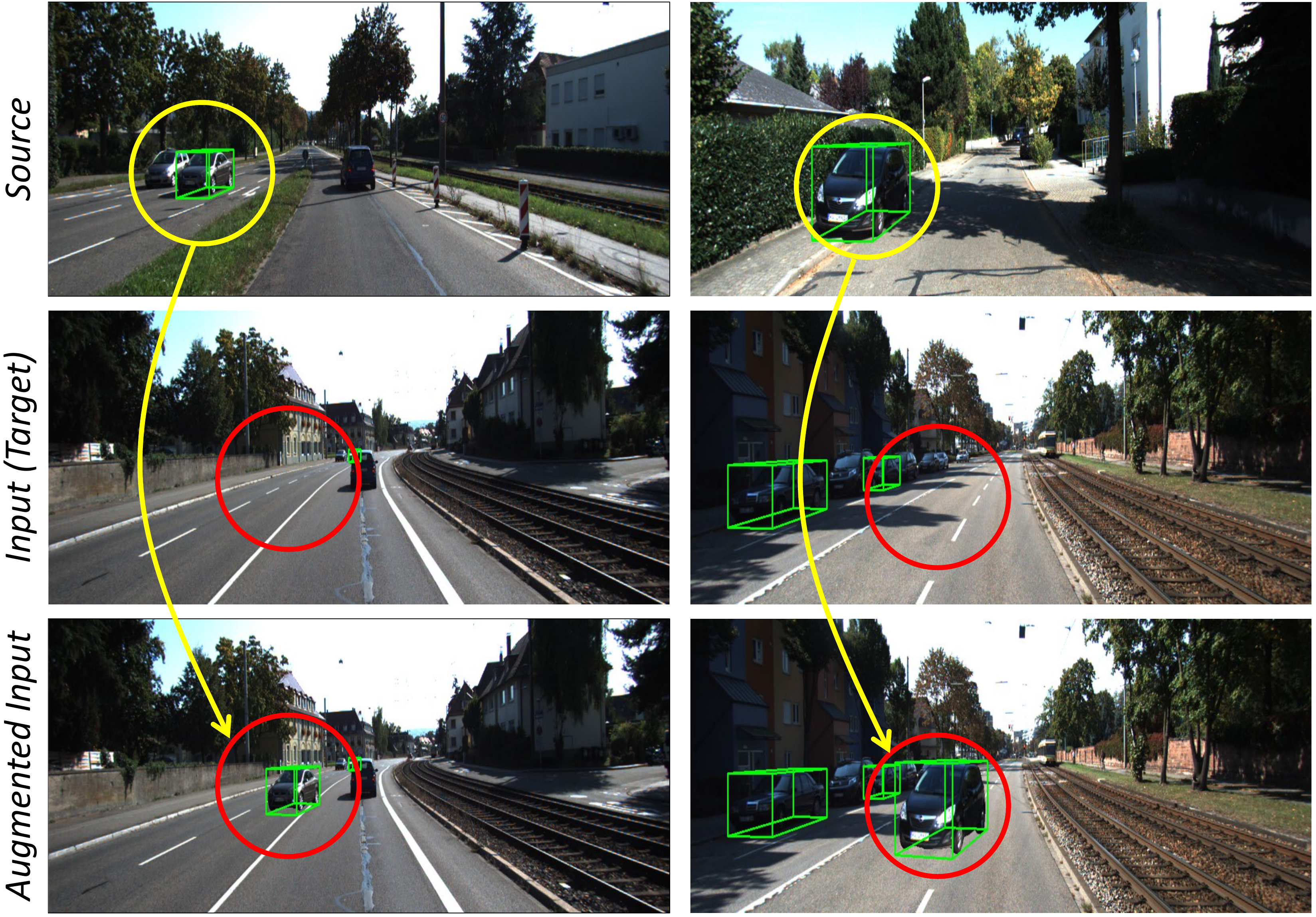}}
\caption{Examples of the RAPA module. Objects segmented from a source image are geometrically transformed and placed onto plausible road regions in the target image. It produces visually realistic and geometrically consistent augmentations.}
\label{fig:patch}
\vspace{-0.2cm}
\end{figure}

\noindent\textbf{Example Results of the RAPA Module.} Figure~\ref{fig:patch} presents qualitative results of the RAPA module. The visualization shows that the RAPA module effectively segments object patches from source images and places them onto valid road regions of the target image while respecting 3D geometric constraints. Unlike conventional copy-paste methods, the RAPA module avoids background artifacts and positions objects at geometrically plausible locations. As a result, the augmented objects blend naturally into the scene, demonstrating that the RAPA module produces realistic training samples with consistent 3D geometry. \\

\noindent\textbf{Limitation.} While our method achieves strong performance, it still inherits the limitations of learning from partially labeled data, where pseudo-labels may still contain errors despite the improvements introduced by the RAPA module and the PBF module. Extending the RAPA module with more diverse placement strategies and strengthening the depth reliability check within the PBF module offer promising directions for further enhancing robustness and reducing the remaining gap to fully supervised models.

\section{Conclusion}
We present a novel framework for monocular 3D object detection under sparse annotation settings. Our approach addresses the challenges of limited supervision through two complementary modules: RAPA module, which enriches training data with geometry-consistent object placement, and PBF module, which selects reliable pseudo-labels by jointly considering depth uncertainty and prototype similarity. Together, these components enable the proposed model to learn more stable and informative 3D features even with significantly reduced annotations. Experiments on the KITTI dataset demonstrate substantial improvements over existing SAOD methods, particularly under highly sparse conditions, highlighting the effectiveness of our sparsely annotated approach for learning reliable monocular 3D representations from limited annotations.

\section*{Acknowledgements}
This work was partly supported by the IITP-ITRC grant funded by the Korea government (MSIT)(IITP-2026-RS-2023-00258649, 25\%) and partly supported by IITP grant funded by the Korea government (MSIT) (IITP-2022-II220078: Explainable Logical Reasoning for Medical Knowledge Generation (25\%), IITP-2023-RS-2023-00266615: Convergence Security Core Talent Training Business Support Program (10\%), No. RS-2022-II220124, Development of Artificial Intelligence Technology for Self-Improving Competency-Aware Learning Capabilities(40\%)).

%% file: sec/X_suppl.tex
\def\maketitlesupplementary{
    \clearpage
    \twocolumn[
        \centering
        {\Large \bf \thetitle \par -- \textit{Supplementary Material} --\\ }
        \vspace{1.5em}
    ]
}

\maketitlesupplementary

\makeatletter
\makeatother

\setcounter{section}{0}
\renewcommand\thesection{\arabic{section}}
\setcounter{table}{0}
\renewcommand{\thetable}{S.\arabic{table}}
\setcounter{figure}{0}
\renewcommand{\thefigure}{S.\arabic{figure}}
\setcounter{equation}{0}
\renewcommand{\theequation}{S.\arabic{equation}}

\begin{algorithm}[t]
\caption{Road-Aware Patch Augmentation (RAPA)}
\label{alg:rapa}
\begin{algorithmic}
\State \textbf{Input:} Source patch $P_s$ with label $l_s = (x_s, y_s, z_s, h, w, l, r_y)$, target image $I_t$, camera extrinsics $[R_s|T_s]$, $[R_t|T_t]$, road mask $M_{\text{road}}$, existing boxes $\mathcal{B}_{\text{existing}}$
\State \textbf{Output:} Augmented image $I_t'$ and label $l_t'$
\State
\setcounter{ALG@line}{0}
\end{algorithmic}
\begin{algorithmic}[1]
\State $(x_t, y_t, z_t)^\top = [R_t|T_t][R_s|T_s]^{-1}(x_s, y_s, z_s)^\top$
\State
\For{$n = 1$ to $N_{\text{max}}$}
    \State Sample $x_{\text{offset}} \in [-\delta, \delta]$
    \State $(x_t', y_t', z_t')^\top = (x_t, y_t, z_t)^\top + (x_{\text{offset}}, 0, 0)^\top$
    \State $\theta_{\text{new}} = \arctan2(x_t', z_t')$
    \State $r_y' = \alpha + \theta_{\text{new}}$
    \State Project $(x_t', y_t', z_t', h, w, l, r_y')$ to 2D box $\mathbf{b}$
    \If{$\frac{\sum_{(i,j) \in \mathbf{b}} M_{\text{road}}(i,j)}{|\mathbf{b}|} < \tau_{\text{road}}$}
        \State \textbf{continue}
    \EndIf
    \If{$\exists \mathbf{b}_e \in \mathcal{B}_{\text{existing}}: \mathrm{IoU}(\mathbf{b}, \mathbf{b}_e) \geq \tau_{\text{overlap}}$}
        \State \textbf{continue}
    \EndIf
    \State \textbf{break}
\EndFor
\State
\State Resize $P_s$ to match $\mathbf{b}$ and paste onto $I_t$ → $I_t'$
\State $l_t' = (x_t', y_t', z_t', h, w, l, r_y')$
\State \Return $(I_t', l_t')$
\end{algorithmic}
\end{algorithm}

\begin{figure}[t]
  \centering
  \includegraphics[width=0.93\linewidth]{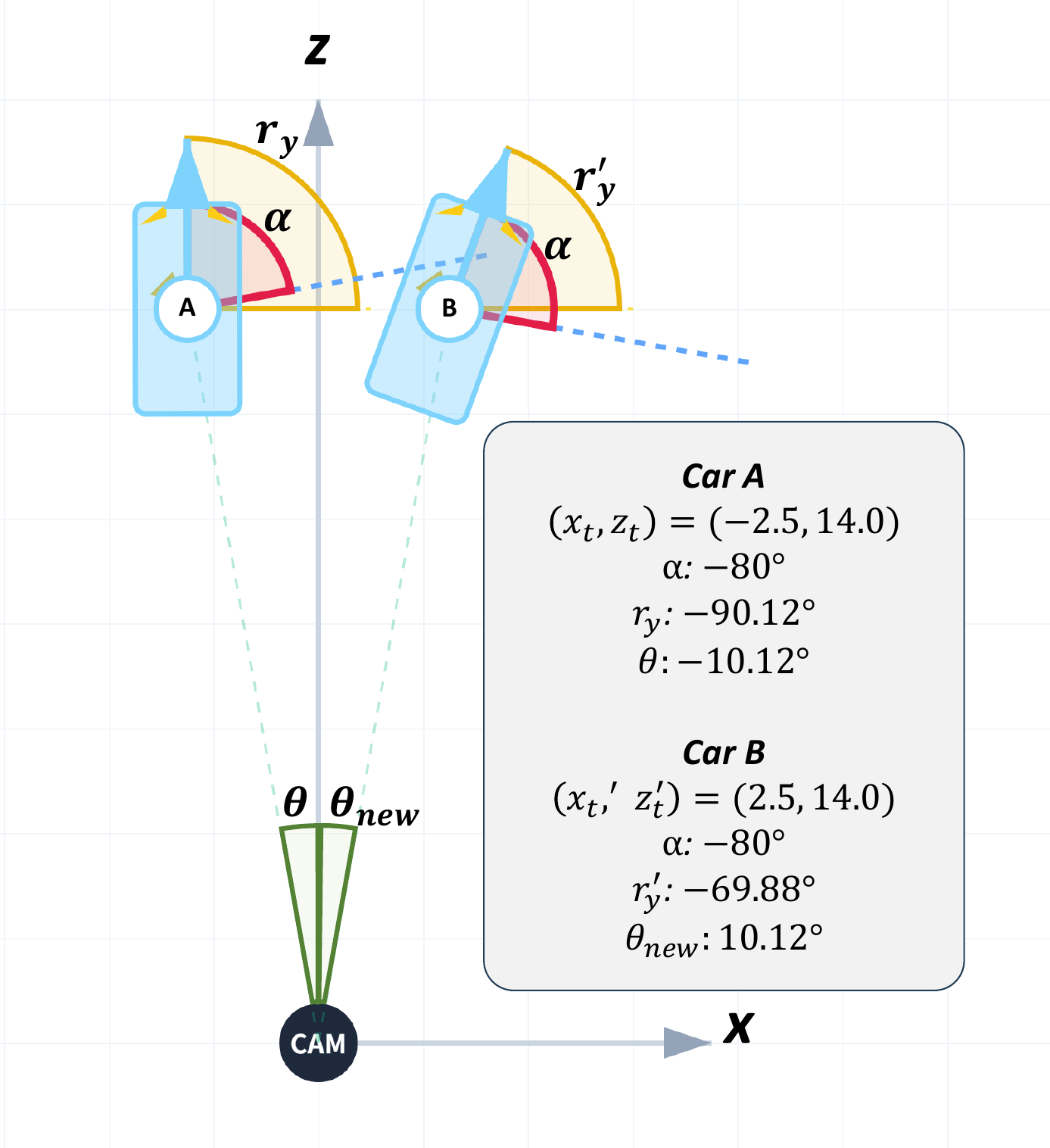}
  \caption{Conceptual visualization of 3D geometric translation of RAPA. The coordinates $(x_t', z_t')$ represent the new 3D position of the object after applying the horizontal translation. Car A is translated to position B while preserving the observation angle $\alpha$ to maintain consistent patch appearance. The global orientation $r_y$ is updated to $r_y'$ based on the new camera viewing angle ($\theta$ to $\theta_{new}$), ensuring 3D geometric consistency.}
  \label{fig:rapa_concept}
\end{figure}

\begin{table*}[t]
  \caption{Detection comparison on the Clear and Foggy KITTI images under the 30\% annotation ratio. We report both $\text{AP}_{3D}$ and $\text{AP}_{BEV}$.}
  \renewcommand{\tabcolsep}{3mm}
  \label{tab:clear_foggy}
  \centering
  \resizebox{0.95\linewidth}{!}{
  \begin{tabular}{@{}c ccc ccc ccc ccc@{}}
    \Xhline{3\arrayrulewidth}
    \rule{0pt}{10pt}
    \multirow{3}{*}{\bf Method} 
    & \multicolumn{6}{c}{\bf Clear Image (30\%)} 
    & \multicolumn{6}{c}{\bf Foggy Image (30\%)} \\
    \cmidrule(lr){2-7} \cmidrule(lr){8-13}
    & \multicolumn{3}{c}{\bf $\text{AP}_{3D}$}
    & \multicolumn{3}{c}{\bf $\text{AP}_{BEV}$}
    & \multicolumn{3}{c}{\bf $\text{AP}_{3D}$}
    & \multicolumn{3}{c}{\bf $\text{AP}_{BEV}$} \\
    \cmidrule(lr){2-4} \cmidrule(lr){5-7}
    \cmidrule(lr){8-10} \cmidrule(lr){11-13}
    & Easy & Mod. & Hard & Easy & Mod. & Hard
    & Easy & Mod. & Hard & Easy & Mod. & Hard \\
    \Xhline{3\arrayrulewidth}
    \rule{0pt}{9pt}Baseline 
      & 11.17 & 8.73 & 7.56 & 17.24 & 13.62 & 11.49 & 11.10 & 7.43 & 5.81 & 16.78 & 11.99 & 9.91 \\
      \cdashline{1-13}
      
    Co-mining
      & 16.01 & 12.62 & 10.38 & \underline{24.81} & 18.31 & 15.28 & 11.22 & 7.91 & 6.20 & 17.04 & 12.09 & 9.76 \\

    SparseDet
      & 16.95 & \underline{13.30} & \underline{10.97} & 24.78 & \underline{19.41} & \underline{16.52} & 11.40 & 7.93 & 6.52 & 17.64 & 12.77 & 10.41 \\

    Calibrated Teacher
      & \underline{17.14} & 12.96 & 10.58 & 24.35 & 18.04 & 15.01 & \underline{11.97} & \underline{8.65} & \underline{7.26} & \underline{18.99} & \underline{13.86} & \underline{11.53} \\
    
    Co-student
      & 15.99 & 12.67 & 10.38 & 24.76 & 18.27 & 15.20 & 11.57 & 7.97 & 6.20 & 17.58 & 12.36 & 9.97 \\ 
      \cdashline{1-13}
    
    \textbf{Proposed}
      & \textbf{21.28} & \textbf{15.60} & \textbf{12.79}
      & \textbf{28.45} & \textbf{20.40} & \textbf{17.04}
      & \textbf{19.11} & \textbf{13.72} & \textbf{10.35}
      & \textbf{28.28} & \textbf{19.49} & \textbf{15.16} \\
    \Xhline{3\arrayrulewidth}
  \end{tabular}
  }
  \vspace{-0.2cm}
\end{table*}

\begin{table*}[t]
    \vspace{0.5cm}
  \caption{Detection results of car category on the KITTI validation set under the lower annotation ratios of 10\%, 20\%, and 30\%. We compare our method with existing SAOD approaches reproduced using their official implementations for fair comparison. \textbf{Bold}/\underline{underline} fonts indicate the best/second-best results.}
  \renewcommand{\tabcolsep}{3.5mm}
  \label{tab:sota_comparison_ultrasparse}
  \centering
  \resizebox{0.94\linewidth}{!}{
  \begin{tabular}{@{}ccccccccccc@{}}
    \Xhline{3\arrayrulewidth}
    \rule{0pt}{10pt}\multirow{2}{*}{\bf Method} & \multicolumn{3}{c}{\bf 10\%} & \multicolumn{3}{c}{\bf 20\%} & \multicolumn{3}{c}{\bf 30\%}\\ \cmidrule(lr){2-4} \cmidrule(lr){5-7} \cmidrule(lr){8-10}
    & \textbf{Easy} & \textbf{Mod.} & \textbf{Hard}
    & \textbf{Easy} & \textbf{Mod.} & \textbf{Hard}
    & \textbf{Easy} & \textbf{Mod.} & \textbf{Hard} \\\hline
        \rule{0pt}{9.0pt}Baseline [\textcolor{cyan}{8}]
      & 1.04 & 1.15 & 0.15
      & 5.38 & 4.70 & 3.70
      & 11.17 & 8.73 & 7.56 \\\cdashline{1-10}
        \rule{0pt}{9.0pt}
        
    Co-mining [\textcolor{cyan}{5}]
      & 0.00 & \underline{2.50} & \underline{2.50}
      & 3.88 & 3.90 & 3.26
      & 16.01 & 12.62 & 10.38 \\

    SparseDet [\textcolor{cyan}{3}]
      & \underline{1.88} & 1.67 & 1.67
      & \underline{8.83} & \underline{7.07} & \underline{5.59}
      & 16.95 & \underline{13.30} & \underline{10.97} \\

    Calibrated Teacher [\textcolor{cyan}{4}]
      & 1.62 & 1.55 & 1.05
      & 6.04 & 4.88 & 4.17
      & \underline{17.14} & 12.96 & 10.58 \\

    Co-student [\textcolor{cyan}{6}]
      & 0.00 & \underline{2.50} & \underline{2.50}
      & 1.25 & 1.66 & 1.66
      & 15.99 & 12.67 & 10.38
     \\\cdashline{1-10}
    \rule{0pt}{9.0pt}\textbf{Proposed Method}
      & \textbf{14.18} & \textbf{10.12} & \textbf{7.58}
      & \textbf{19.48} & \textbf{13.78} & \textbf{10.93}
      & \textbf{21.28} & \textbf{15.60} & \textbf{12.79} \\
    \Xhline{3\arrayrulewidth}
  \end{tabular}
  }
  \vspace{-0.2cm}
\end{table*}

\section{Additional Implementation Details}

In Table~3 of the main paper, we set the depth reliability threshold to $\tau_{\text{depth}} = 0.7$ for the setting without RAPA, while all other experiments use $\tau_{\text{depth}} = 1.0$. This adjustment is necessary because Table~3 of the main paper evaluates the model without RAPA, resulting in a different depth–uncertainty distribution from the full framework. Since RAPA alleviates geometric ambiguity by enforcing more consistent object appearances, the model produces more reliable depth estimates when RAPA is applied. To ensure a fair and controlled comparison in the no-RAPA ablation, we lower $\tau_{\text{depth}}$ accordingly. \\

\noindent\textbf{Additional Implementation Details of RAPA.}
During offline patch extraction, we apply strict filtering to ensure high-quality patches. From the sparse ground-truth annotations, we select only objects with truncation level~0 and occlusion level~0 to ensure full visibility without boundary cuts. We also enforce depth constraints of $2.0 \le z < 65.0$ meters, as MonoDETR [\textcolor{cyan}{8}] and MonoDGP [\textcolor{cyan}{2}] internally filter out labels outside this range; this prevents augmented patches from appearing without valid labels.

For segmentation, SAM is prompted with 2D bounding boxes to extract clean car-only patches with transparent backgrounds, saved as RGBA images. We additionally generate road segmentation masks $M_{\text{road}}$ for each training scene using SAM with manually provided point prompts on drivable areas. Each mask is a binary image where non-zero pixels denote road regions. These masks are created once before training and cached for fast lookup.

For valid placement search (detailed in Algorithm~\ref{alg:rapa}), we set the horizontal search range to $\delta = 5.0$ meters and uniformly sample $m = 10$ candidate offsets, exploring positions within $[-5.0, 5.0]$ meters from the transformed 3D location. Each candidate must satisfy two constraints.
(1) A road-overlap threshold of $\tau_{\text{road}} = 0.7$, requiring at least 70\% of the projected 2D box to lie on drivable regions.
(2) An overlap threshold of $\tau_{\text{overlap}} = 0.1$, limiting IoU with existing labeled objects to at most 10\% to avoid unrealistic occlusions.

Once a valid placement is found, we update the rotation by recomputing the camera-viewing angle to preserve geometric consistency. This viewpoint-driven rotation adjustment is shown in Figure~\ref{fig:rapa_concept}, where the object orientation changes depending on its horizontal position relative to the camera.

We apply RAPA consistently across the entire training pipeline, including the teacher initialization stage (195 epochs). Patch-augmented images are regenerated at every epoch. To avoid duplicating objects within the same scene, we exclude patches originating from the target scene. For each image, we set the maximum number of placement trials to $N_{\text{max}} = 40$, meaning the algorithm attempts up to 40 candidate locations to find a valid road-consistent placement before moving to the next sample.

\section{Additional Experiments}
\noindent\textbf{Additional BEV Results.}
We report both $AP_{3D}$ and $AP_{BEV}$ results under 30\% annotation settings for clear and foggy images in Table~\ref{tab:clear_foggy}. Our method consistently outperforms the existing Sparsely-Annotated Object Detection (SAOD) methods in both metrics. \\

\noindent\textbf{Results under Extremely Sparse Annotation Settings.} We further evaluate our method under extremely sparse annotation settings of 10\% and 20\% in Table~\ref{tab:sota_comparison_ultrasparse}. Our method achieves 10.12 AP (Moderate) at 10\% annotation, significantly outperforming the best baseline SparseDet (1.67 $AP_{3D}$) by +8.45. This large improvement demonstrates that RAPA effectively leverages the limited training data, while PBF provides more reliable pseudo-label selection than confidence-based approaches. At 20\% annotation, our method maintains strong performance with 13.78 AP (Moderate), showing +6.71 improvement over SparseDet. The results validate that our approach is particularly effective in extreme low-data regimes. \\

\noindent\textbf{PBF Module Ablation Study.} Our Prototype-Based Filtering (PBF) module uses two scoring functions: $S_{\text{proto}}$ for prototype similarity and $S_{\text{depth}}$ for depth uncertainty. We ablate these components with and without Road-Aware Patch Augmentation (RAPA) in Table~\ref{tab:ablation_combined}. The results show that both scoring functions contribute to performance gain compared to the baseline, and their effectiveness is further amplified when combined with RAPA. \\

\noindent\textbf{Hyperparameter Sensitivity.}
Table~\ref{tab:hyperparam_kitti} shows that our method is relatively robust to variations in $\tau_{\text{depth}}$ and $\tau_{\text{proto}}$, consistently outperforming existing methods across different settings. However, when $\tau_{\text{proto}}$ is set to an extreme value (\textit{i.e.}, 
$\tau_{\text{proto}}\geq0.95$), overly strict filtering results in too few pseudo-labels being accepted, reducing the effective supervision and leading to a slight performance drop. \\

\begin{table}[t!]
\vspace{0.1cm}
  \caption{Ablation study of PBF modules on the KITTI validation set for the car category under 30\% annotation ratio.}
  \renewcommand{\tabcolsep}{3mm}
  \label{tab:ablation_combined}
  \centering
  \resizebox{0.9\linewidth}{!}{
  \begin{tabular}{cccccc}
    \Xhline{3\arrayrulewidth}
    \textbf{RAPA} &
    $\boldsymbol{S_{\text{proto}}}$ &
    $\boldsymbol{S_{\text{depth}}}$ &
    \textbf{Easy} & \textbf{Mod.} & \textbf{Hard} \\\hline

    - & - & - & 11.17 & 8.73 & 7.56 \\
    \cdashline{1-6}

    \rule{0pt}{9.0pt}
    - & \cmark & - & 16.07 & 12.26 & 9.90 \\
    - & \cmark & \cmark & \textbf{16.49} & \textbf{12.65} & \textbf{10.32} \\

    \cdashline{1-6}
    \cmark & \cmark & - & 18.73 & 14.85 & 12.13 \\
    \cmark & \cmark & \cmark & \textbf{21.28} & \textbf{15.60} & \textbf{12.79} \\

    \Xhline{3\arrayrulewidth}
  \end{tabular}
  }
  \vspace{-0.2cm}
\end{table}

\begin{table}[t!]
\vspace{0.1cm}
\caption{Hyperparameter sensitivity analysis on KITTI (30\%).}
\renewcommand{\tabcolsep}{3mm}
\centering
\label{tab:hyperparam_kitti}
\begin{tabular}{ccccc}
\Xhline{3\arrayrulewidth}
$\boldsymbol{\tau_{\text{depth}}}$ &
$\boldsymbol{\tau_{\text{proto}}}$ &
\textbf{Easy} & \textbf{Mod.} & \textbf{Hard} \\ \hline

0.8 &  & 21.75 & 15.45 & 12.48 \\
1.0 & 0.85 & 21.28 & 15.60 & 12.79 \\
1.2 &  & 22.12 & 15.47 & 12.32 \\

\cdashline{1-5}

 & 0.75 & 21.56 & 15.57 & 12.62 \\
1.0 & 0.85 & 21.28 & 15.60 & 12.79 \\
 & 0.95 & 20.46 & 14.58 & 11.80 \\

\Xhline{3\arrayrulewidth}
\end{tabular}
\vspace{-0.2cm}
\end{table}

\begin{figure*}[t]
  \centering
  \includegraphics[width=\linewidth]{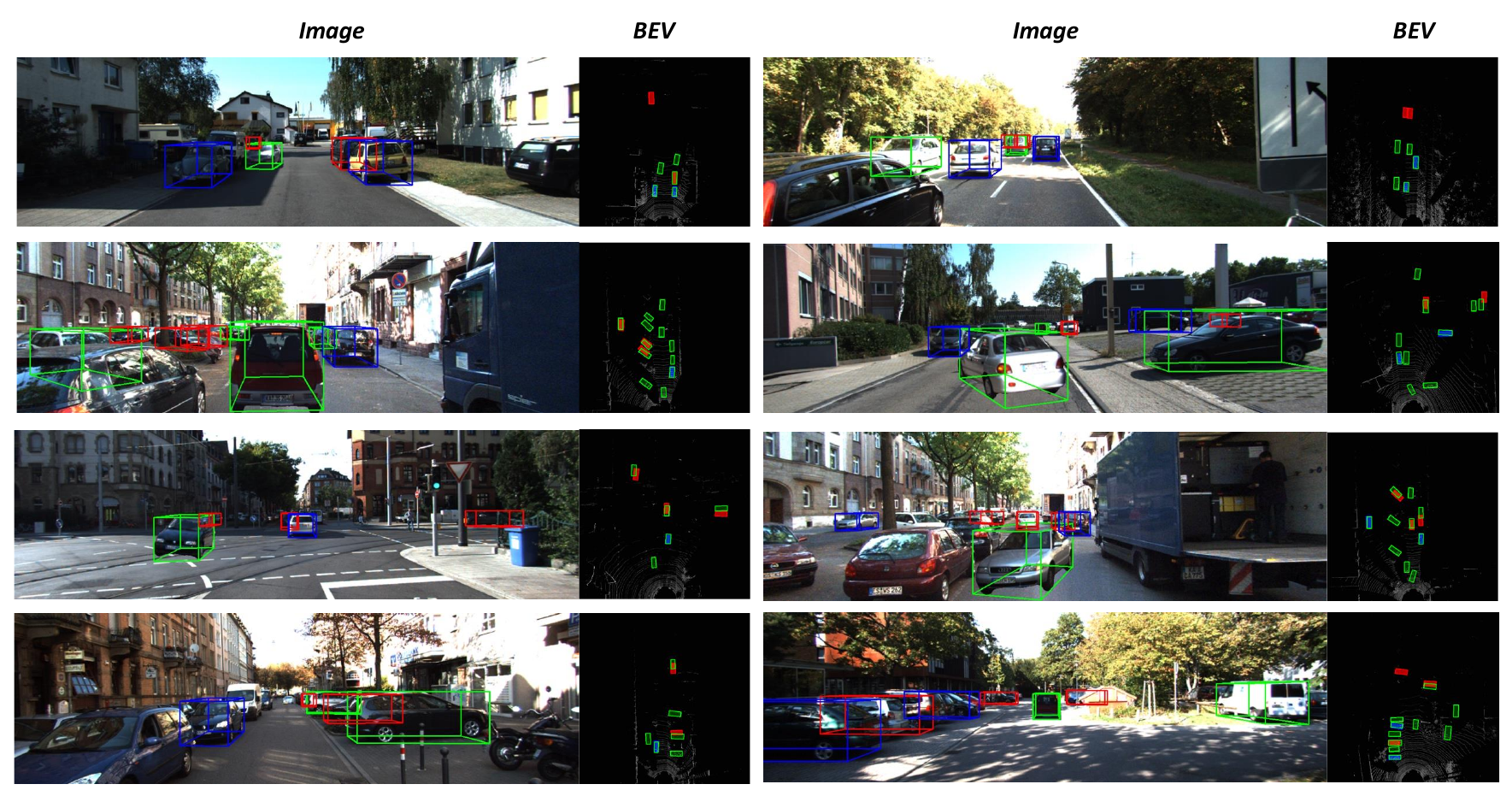}
  \caption{
Visualization of PBF filtering results. 
In the image view, green boxes indicate the sparsely annotated ground-truth labels, while blue boxes denote the selected pseudo-labels and red boxes indicate predictions filtered out by PBF. 
In the BEV view (right), green boxes represent the original full ground-truth annotations, and blue boxes correspond to the retained pseudo-labels. 
The BEV visualization highlights how well the selected pseudo-labels align with the original ground-truth geometry, demonstrating the effectiveness of PBF in filtering out predictions.
}
  \label{fig:pbf_filtering}
\end{figure*}

\noindent\textbf{Computational Complexity and Overhead.}
Our framework introduces additional training cost compared with the baseline due to the teacher--student architecture. However, such a design is commonly adopted in sparse annotation learning frameworks and is used here to ensure fair comparison with prior SAOD methods. Compared with Co-Student [\textcolor{cyan}{5}], our method shows more efficient computation, requiring 1.92 s/iter versus 2.58 s/iter. In addition, the memory overhead is modest, increasing GPU memory usage from 20.45 GB to 21.87 GB (+1.42 GB, 6.9\%). The prototype module itself introduces negligible computational overhead, requiring only 0.38 ms/iter and 31 MB of memory. \\

\begin{figure}[t]
\centering
\includegraphics[width=\linewidth]{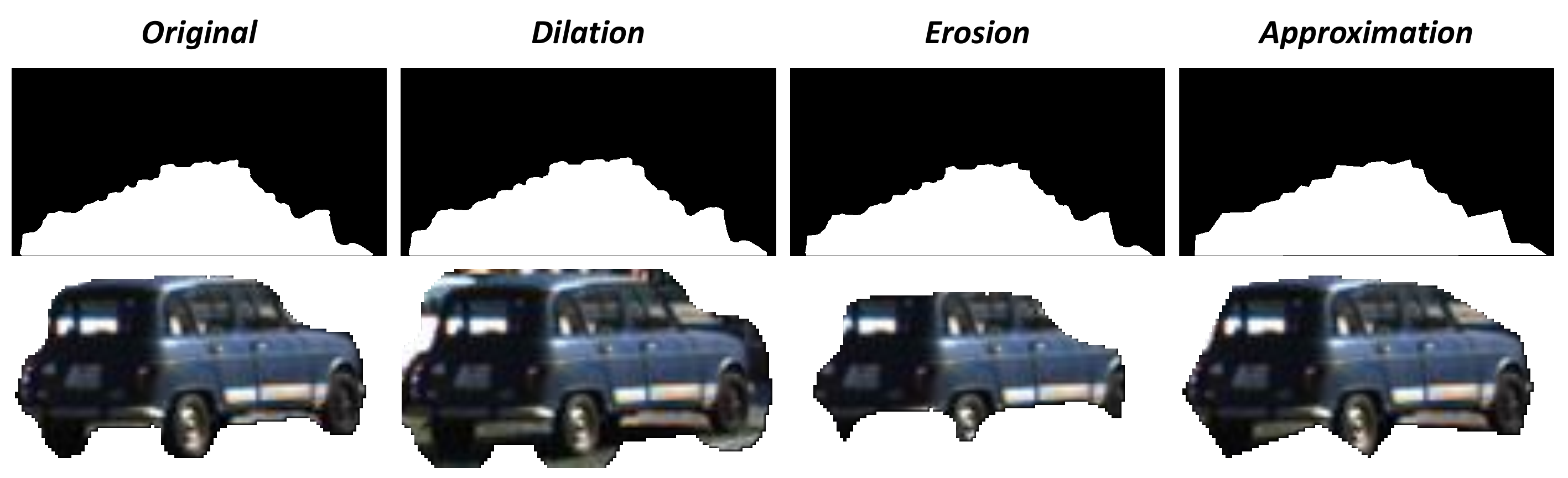}
\caption{Examples of simulated segmentation errors: dilation, erosion, and boundary approximation.}
\label{fig:figuire_noise.pdf}
\end{figure}

\noindent\textbf{Robustness against segmentation noise.} To evaluate whether our method depends on precise segmentation quality, we simulate common segmentation errors following prior protocols [\textcolor{cyan}{1}], by injecting boundary perturbations to both road and object masks. Specifically, we apply three types of noise: dilation, erosion (5px), and polygonal boundary approximation, as illustrated in Figure~\ref{fig:figuire_noise.pdf}. Table~\ref{tab:sam_robust} reports the results under these corrupted masks on KITTI with 30\% annotations. Despite mask distortions, our method maintains stable performance across all settings. This robustness arises because segmentation masks are used only for coarse region extraction and placement zone identification. The final object placement and geometry are governed by the 3D geometric constraints in Eq.~(1--2, 5--6), which remain unaffected by small boundary inaccuracies. Consequently, approximate masks are sufficient, indicating that our method does not rely on precise segmentation quality or SAM-specific accuracy.

\noindent\textbf{Comparison with other Patch Augmentation Methods.} We compare RAPA with existing patch augmentation methods Mix-Teaching [\textcolor{cyan}{7}] and CMAug [\textcolor{cyan}{9}] in Table~\ref{tab:patch_adding}. To isolate the contribution of augmentation from pseudo-label filtering, we report results both with and without our PBF module. RAPA outperforms both baselines even without PBF (14.51 vs. 13.30 and 13.26 $AP_{3D}$ on Moderate), demonstrating the effectiveness of our road-aware placement strategy and 3D geometric consistency. When combined with PBF, RAPA achieves the best performance of 15.60 $AP_{3D}$ on Moderate.

\section{Additional Visualization Results}
\noindent\textbf{Effect of the PBF Module.}
Figure~\ref{fig:pbf_filtering} visualizes the filtering behavior of PBF.
In the image view, green boxes represent the sparsely annotated ground-truth labels, while blue boxes denote the selected pseudo-labels and red boxes indicate predictions discarded by PBF due to low prototype similarity or high depth uncertainty.
In the BEV view, green boxes correspond to the full ground-truth annotations (not the sparse subset), enabling a direct comparison between the retained pseudo-labels (blue) and the true object geometry.
To provide a broader set of qualitative examples, we additionally include predictions with confidence scores of at least 0.2.
As shown in the figure, the discarded predictions often exhibit incorrect depth estimates, whereas the selected pseudo-labels align well with the original ground truth, demonstrating the effectiveness of PBF in distinguishing reliable predictions from erroneous ones.

\begin{table}[t]
    \vspace{0.075cm}
    \caption{Comparison with other patch augmentation methods used in monocular 3D object detection tasks, on the KITTI validation set under the 30\% annotation ratio. Results are reported with and without PBF to show the isolated contribution of each augmentation approach.}
    \renewcommand{\tabcolsep}{4mm}
    \label{tab:patch_adding}
    \centering
    \resizebox{\linewidth}{!}{
    \begin{tabular}{l ccc}
        \Xhline{3\arrayrulewidth}
        \textbf{Method}\rule{0pt}{9.5pt} & \textbf{Easy} & \textbf{Mod.} & \textbf{Hard} \\\hline
        \rule{0pt}{9.0pt}Mix-Teaching [\textcolor{cyan}{7}] 
            & 15.42 & 13.26 & 11.17 \\
        CMAug [\textcolor{cyan}{9}]
            & 17.15 & 13.30 & 11.44 \\
        \textbf{RAPA}
            & \textbf{20.31} & \textbf{14.51} & \textbf{11.72} \\\cdashline{1-4}
        \rule{0pt}{9.0pt}Mix-Teaching + PBF 
            & 18.96 & 14.77 & 12.22 \\
        CMAug + PBF 
            & 18.93 & 14.79 & 12.61 \\
        \textbf{RAPA + PBF} 
            & \textbf{21.28} & \textbf{15.60} & \textbf{12.79} \\
        \Xhline{3\arrayrulewidth}
    \end{tabular}
    }
    \vspace{-0.2cm}
\end{table}

\begin{table}[h]
\centering
\small
\caption{Robustness to segmentation noise on KITTI with 30\% annotations.}
\label{tab:sam_robust}
\begin{tabular}{lccc}
\toprule
Method & Easy & Mod. & Hard \\
\midrule
Baseline (No Aug) & 11.17 & 8.73 & 7.56 \\
Scale (Dilation) & 18.47 & 14.07 & 11.38 \\
Scale (Erosion) & 18.04 & 13.86 & 11.19 \\
Boundary Approx. & 20.04 & 14.30 & 11.54 \\
SAM (Ours) & \textbf{20.31} & \textbf{14.51} & \textbf{11.72} \\
\bottomrule
\end{tabular}
\end{table}

\begin{figure*}[t]
  \centering
  \includegraphics[width=\linewidth]{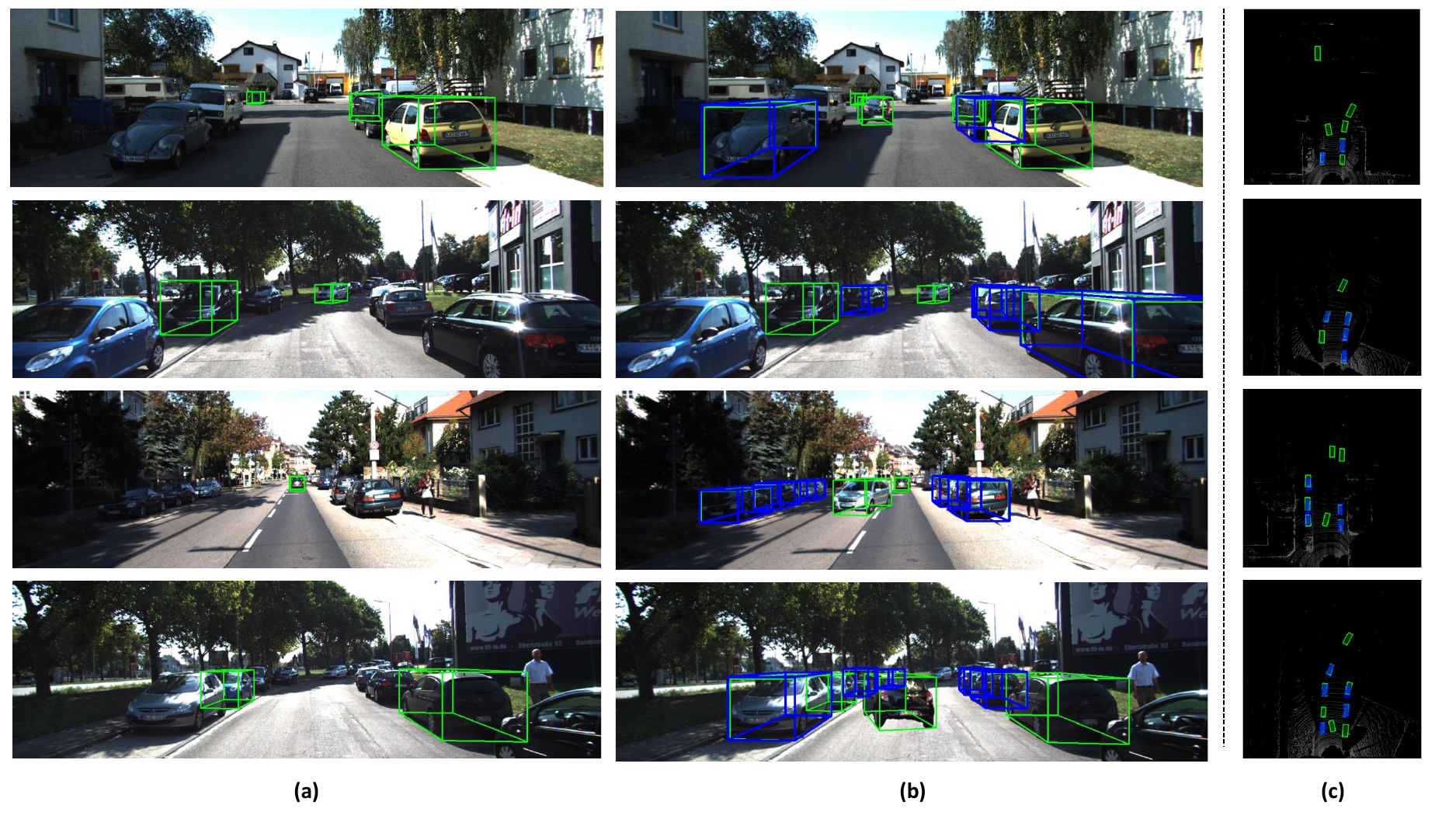}
   \caption{
Visualization of pseudo-labels added during training. 
(a) shows the sparsely annotated ground-truth labels. 
(b) presents the combined set of ground-truth and newly added pseudo-labels, where blue boxes denote the newly added pseudo-labels and green boxes represent all available labels (sparse ground-truth + accumulated pseudo-labels). 
(c) visualizes the BEV alignment between the newly added pseudo-labels (blue) and the original full ground-truth annotations (green), illustrating that the generated pseudo-labels are geometrically consistent with the true object locations.}
  \label{fig:pseudo_added_example}
\end{figure*}

\noindent\textbf{Visualization of the GT Bank Updating Process.}
Figure~\ref{fig:pseudo_added_example} provides additional details on how pseudo-labels are accumulated throughout training.
In (a), only the sparsely annotated ground-truth boxes are available.
As training progresses, PBF selects reliable predictions based on prototype similarity and depth uncertainty, and (b) shows the combined set of sparse ground truth and newly added pseudo-labels, where the latter are highlighted in blue.
The green boxes in (b) indicate the current GT Bank, which contains both the original sparse annotations and the pseudo-labels accumulated from previous iterations.

To assess the geometric correctness of the added pseudo-labels, (c) compares them with the original full ground-truth annotations in the BEV view: green boxes represent the original ground truth, while blue boxes denote the pseudo-labels added to the GT Bank.
The strong alignment between the two sets demonstrates that PBF effectively selects and stores only geometrically reliable predictions in the GT Bank. \\

\begin{figure*}[t]
  \centering
  \includegraphics[width=\linewidth]{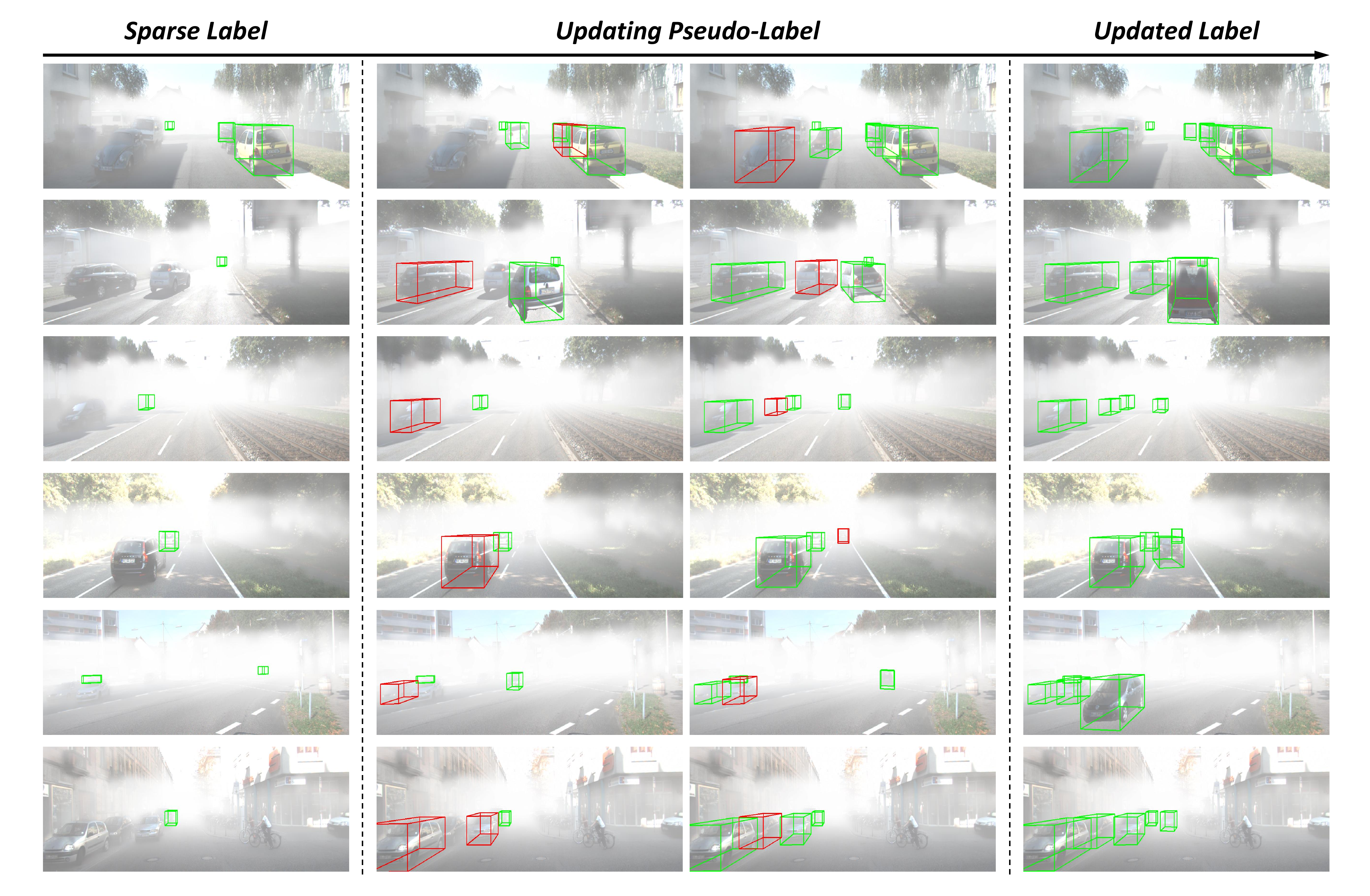}
  \caption{Progression of pseudo-labels selected by the proposed PBF module for GT Bank enrichment on foggy images. Green boxes denote sparse ground truths and previously accumulated pseudo-labels, while red boxes indicate high-quality pseudo-labels newly selected at the current step. The consistent selection of geometrically and semantically reliable pseudo-labels highlights the effectiveness of the PBF module.}
  \label{fig:foggy_progression}
\end{figure*}

\noindent\textbf{Example of Pseudo-labeling Progression in Foggy Images.} We visualize the progression of our method on foggy images in Figure~\ref{fig:foggy_progression}. Similar to clear images shown in the main paper (Figure 3), our framework progressively generates high-quality pseudo-labels even under adverse weather conditions. As training progresses, low-quality predictions are filtered out by PBF, while reliable predictions that satisfy both prototype similarity and depth uncertainty criteria are retained as pseudo-labels (shown in red). The results demonstrate that our method maintains robust pseudo-label generation capabilities on foggy images, validating the effectiveness of feature-level filtering over confidence-based approaches in challenging conditions. \\

\begin{figure*}[t]
  \centering
  \includegraphics[width=\linewidth]{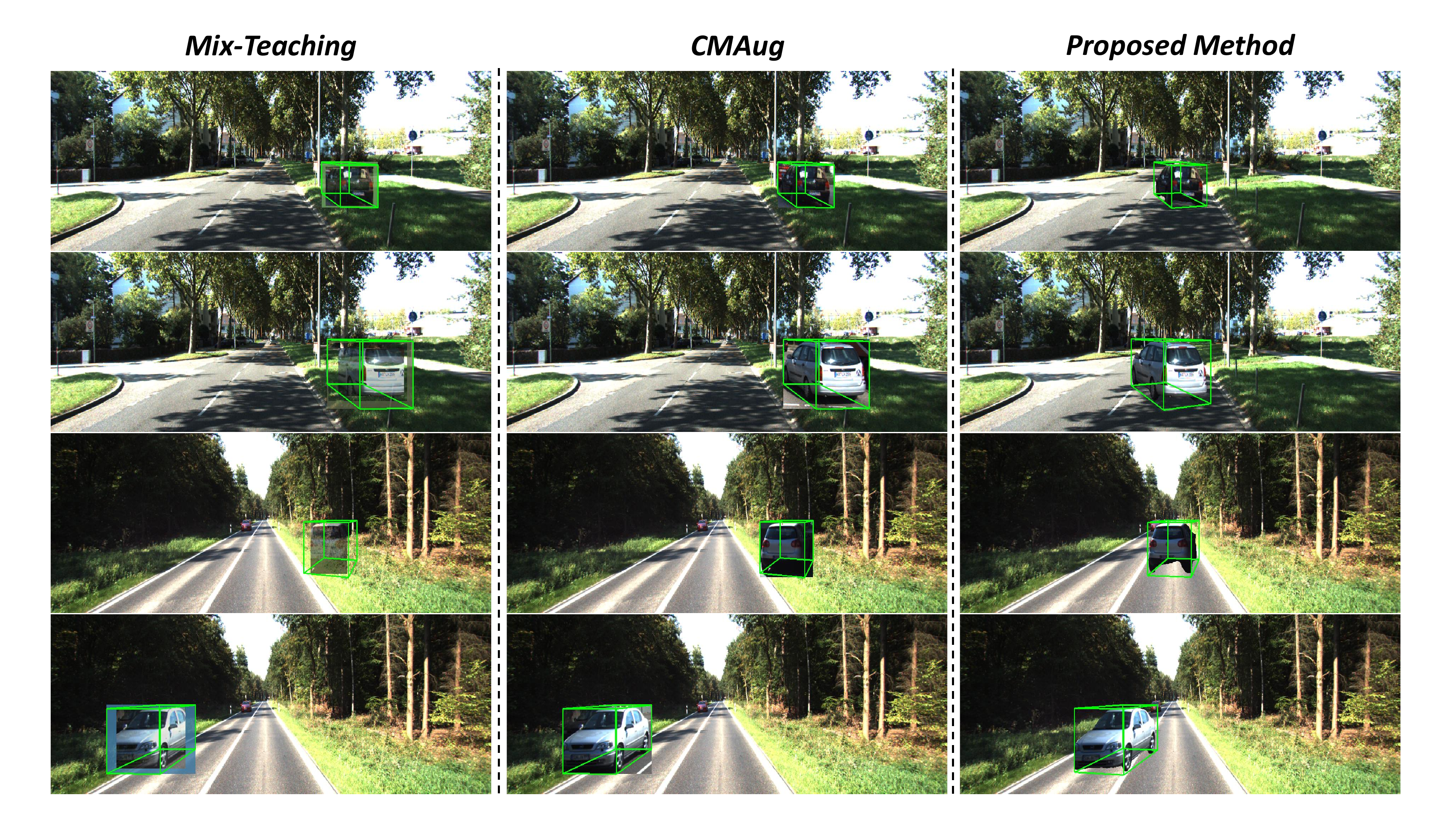}
  \caption{Qualitative comparison of augmentation methods. Proposed method (RAPA) produces realistic augmentations by placing objects at valid locations with geometrically consistent 3D bounding boxes.}
  \label{fig:augmentation_comparison}
\end{figure*}

\noindent\textbf{Visual Comparison with Other Patch Augmentation Methods.} We visually compare RAPA with other augmentation methods in Figure~\ref{fig:augmentation_comparison}. Mix-Teaching [\textcolor{cyan}{7}] applies random border cut, color-padding, and mix-up for patch-level augmentation, but maintains objects at their original locations. CMAug [\textcolor{cyan}{9}] considers 3D transformations when placing patches, but uses classic copy-paste augmentation that includes background context, creating unrealistic composite images. In contrast, RAPA extracts clean object patches using SAM and places them at geometrically valid road locations. This generates more realistic training samples while preserving 3D geometric consistency.

\section*{References}
\small
\begin{list}{[\arabic{enumi}]}{
    \usecounter{enumi}
    \setlength{\leftmargin}{1.8em}
    \setlength{\itemindent}{0em}
    \setlength{\itemsep}{0.2em}
    \setlength{\parsep}{0em}
    \setlength{\topsep}{0em}
    \setlength{\labelsep}{0.5em}
    \setlength{\labelwidth}{1.3em}
}
\item Bowen Cheng, Ross Girshick, Piotr Doll{\'a}r, Alexander C Berg, and Alexander Kirillov.
Boundary IoU: Improving object-centric image segmentation evaluation.
In \textit{CVPR}, 2021.

\item Fanqi Pu, Yifan Wang, Jiru Deng, and Wenming Yang.
Monodgp: Monocular 3D object detection with decoupled-query and geometry-error priors.
In \textit{CVPR}, 2025.

\item Saksham Suri, Saketh Rambhatla, Rama Chellappa, and Abhinav Shrivastava.
Sparsedet: Improving sparsely annotated object detection with pseudo-positive mining.
In \textit{ICCV}, 2023.

\item Haohan Wang, Liang Liu, Boshen Zhang, Jiangning Zhang, Wuhao Zhang, Zhenye Gan, Yabiao Wang, Chengjie Wang, and Haoqian Wang.
Calibrated teacher for sparsely annotated object detection.
In \textit{AAAI}, 2023.

\item Tiancai Wang, Tong Yang, Jiale Cao, and Xiangyu Zhang.
Co-mining: Self-supervised learning for sparsely annotated object detection.
In \textit{AAAI}, 2021.

\item Lianjun Wu, Jiangxiao Han, Zengqiang Zheng, and Xinggang Wang.
Co-Student: Collaborating Strong and Weak Students for Sparsely Annotated Object Detection.
In \textit{ECCV}, 2024.

\item Lei Yang, Xinyu Zhang, Jun Li, Li Wang, Minghan Zhu, Chuang Zhang, and Huaping Liu.
Mix-teaching: A simple, unified and effective semi-supervised learning framework for monocular 3D object detection.
In \textit{TCSVT}, 2023.

\item Renrui Zhang, Han Qiu, Tai Wang, Ziyu Guo, Ziteng Cui, Yu Qiao, Hongsheng Li, and Peng Gao.
MonoDETR: Depth-guided transformer for monocular 3D object detection.
In \textit{ICCV}, 2023.

\item Weijia Zhang, Dongnan Liu, Chao Ma, and Weidong Cai.
Alleviating foreground sparsity for semi-supervised monocular 3D object detection.
In \textit{WACV}, 2024.
\end{list}

